\documentclass{article} % For LaTeX2e
\usepackage{PRIMEarxiv}
\usepackage[utf8]{inputenc}
\usepackage[T1]{fontenc}
\usepackage{natbib}
\setcitestyle{authoryear,round,citesep={;},aysep={,},yysep={;}}

\usepackage{amsmath,amsfonts,bm}

\def\eqref#1{equation~\ref{#1}}
\def\1{\bm{1}}

\DeclareMathAlphabet{\mathsfit}{\encodingdefault}{\sfdefault}{m}{sl}
\SetMathAlphabet{\mathsfit}{bold}{\encodingdefault}{\sfdefault}{bx}{n}

\usepackage{hyperref}
\usepackage{url}
\usepackage{graphicx}
\usepackage{subcaption}
\usepackage{booktabs}
\usepackage{placeins}
\usepackage{float}
\usepackage{multirow}

\title{LiftGCN: Efficient Energy-Preserving Graph Learning via Joukowski Spectral Lifting for Finite Element Stress Prediction}

\author{
Chen Zeng\textsuperscript{1}, Qiao Wang\textsuperscript{1,2}\thanks{Corresponding author. ORCID: 0000-0002-5271-0472}
\\
\textsuperscript{1}School of Information Science and Engineering, Southeast University \\
\textsuperscript{2}School of Economics and Management, Southeast University \\
Nanjing, China \\
\texttt{\{chenzeng,qiaowang\}@seu.edu.cn}
}

\begin{document}

\maketitle

\begin{abstract}
Finite element stress fields often exhibit strong local non-smoothness, where stress concentrations near holes, notches, and loading regions induce sharp spatial gradients and high-frequency graph components. Although graph neural networks naturally operate on irregular finite element meshes, conventional message passing is inherently smoothing and progressively attenuates such high-frequency information. Unitary propagation alleviates this problem by preserving spectral magnitudes, but typically relies on matrix functions and high-order approximations with $O(Ked)$ propagation complexity. We propose LiftGCN, an efficient spectrally stable graph network based on Joukowski spectral lifting. LiftGCN maps the real spectrum of a normalized graph operator onto the unit circle through the Joukowski relation and realizes the resulting spectral transformation as a simple second-order recurrence, avoiding matrix exponentials, eigendecomposition, and high-order polynomial truncation. We show that the linear Joukowski backbone has unit-modulus characteristic roots and admits an energy-preserving structure under a positive-definite metric, preventing exponential attenuation of graph-frequency components with depth. Each layer requires only one sparse neighborhood aggregation, yielding $O(ed)$ propagation complexity, while lightweight local nonlinear residuals provide expressive feature transformations. Experiments on finite element stress prediction demonstrate that LiftGCN achieves competitive overall accuracy while improving reconstruction of stress concentrations and local high-gradient structures with substantially reduced computational cost. Our codee is available at \url{https://github.com/ChenZeng001/LiftGCN}.
\end{abstract}

\section{Introduction}

\begin{figure}[t]
  \centering
  \includegraphics[width=\textwidth]
  {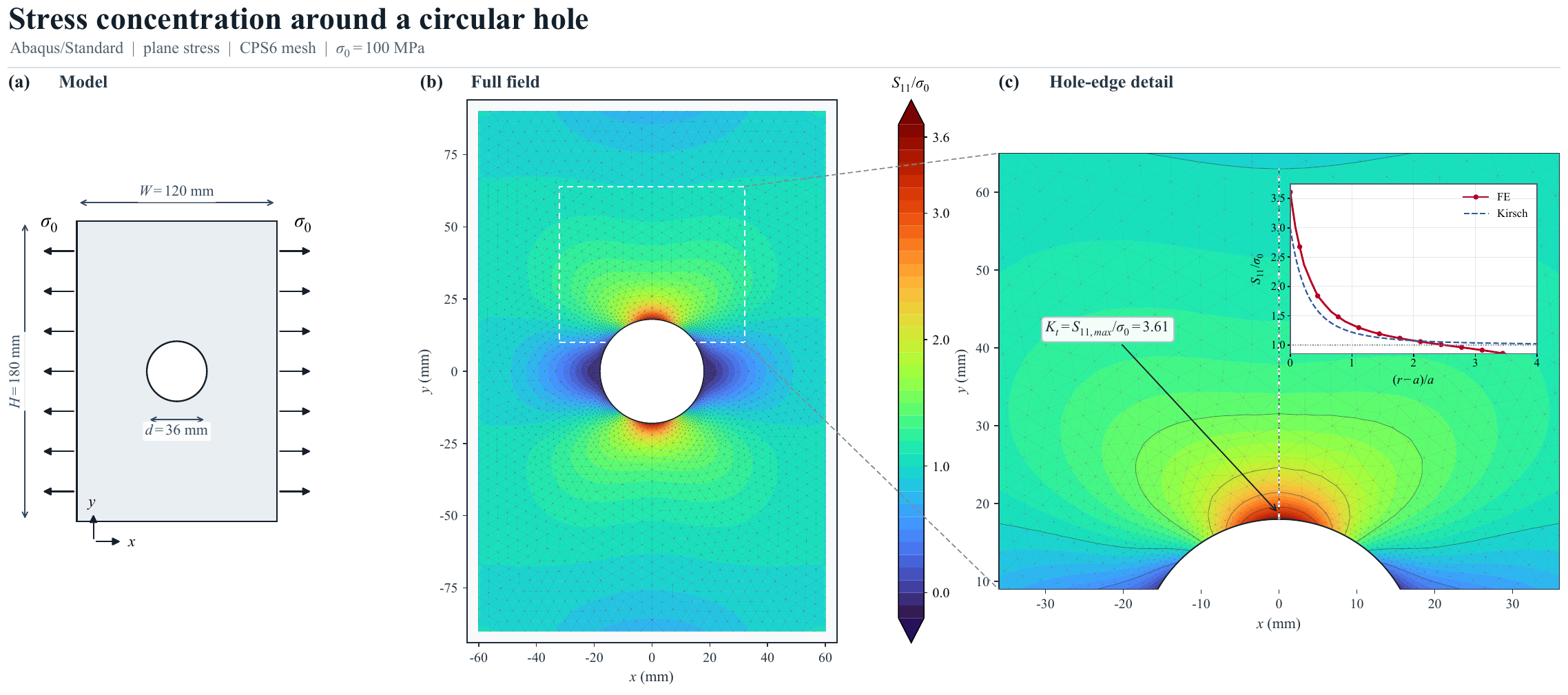}
  \caption{Stress concentration around a circular hole: geometry and loading, the finite element stress field, and local stress decay compared with the Kirsch solution.}
  \label{fig:plate-hole-stress}
\end{figure}

Finite element analysis (FEA) is widely used to predict mechanical responses, but repeated high-resolution simulations can be expensive~\citep{bathe1996finite,zienkiewicz2013finite,jayasinghe2025review}. This cost motivates learned surrogates for design optimization and structural evaluation~\citep{li2021development,giacomini2026surrogates,liang2018deep}. Stress prediction is particularly demanding: a field may be smooth over much of the domain yet contain sharp concentrations near holes, notches, interfaces, and loading regions~\citep{sun2024end}. These local responses are important for assessing structural failure, so a useful surrogate must recover both the overall field and its steep spatial variations~\citep{ZHU2022103513,2020gradient}. Figure~\ref{fig:plate-hole-stress} illustrates this challenge.

Graph neural networks (GNNs) operate directly on irregular finite element meshes, using connectivity to exchange information between nodes~\citep{GNN,GNNreview}. MeshGraphNets and related models have demonstrated their value for physical simulation and structural prediction~\citep{MeshGraphNets,maurizi2022predicting}. However, repeated neighborhood aggregation tends to smooth node features and attenuate graph-frequency components~\citep{li2018deeper,hoang2021revisiting,roth2024rank}. In stress fields, these components can encode mechanically meaningful concentrations and gradients. Smoothing them can therefore remove precisely the local details that a surrogate needs to retain.

Several approaches seek less dissipative propagation. Graph-Coupled Oscillator Networks (GraphCON) introduce second-order dynamics, using inertia to reduce the collapse of deep representations~\citep{GraphCON}. Wave-driven GNNs similarly replace diffusion with oscillatory information transport~\citep{Wave-driven-GNN}. These methods show how retaining a previous state can change the behavior of repeated graph aggregation and help preserve information over depth.

A complementary line of work controls propagation in the spectral domain. CayleyNets use rational filters to obtain flexible frequency responses~\citep{Cayleynet}. Unitary Graph Convolution maps a symmetric graph operator to the complex unit circle, preserving the magnitude of every spectral component~\citep{UniGCN}. This suggests a useful design principle for stress prediction: graph frequencies should evolve through phase while retaining the information needed to reconstruct sharp spatial variations.

The practical difficulty is evaluating the propagation operator. Matrix exponentials on sparse graphs are typically implemented through Taylor, Chebyshev, or related approximations. A $K$-term approximation requires repeated sparse operations and incurs $O(Ked)$ propagation cost, where $e$ is the number of edges and $d$ is the hidden width~\citep{UniGCN}. Obtaining unit-modulus dynamics with a single neighborhood aggregation would make this principle more accessible to large finite element meshes.

We propose LiftGCN to obtain unit-modulus spectral dynamics with the locality and cost of ordinary message passing. The Joukowski relation~\citep{Joukowski} converts a real graph eigenvalue into a pair of unit-circle modes. We realize this mapping through a real-valued second-order recurrence whose linear backbone preserves a positive-definite quadratic energy. Lightweight node-wise residuals supply nonlinear feature transformations, and each layer needs only one sparse neighborhood aggregation. This yields $O(ed)$ graph propagation without evaluating a matrix function.

The remainder of this paper is organized as follows. Section~\ref{sec:background} introduces finite element stress fields and reviews diffusive and unitary graph propagation. 
Section~\ref{sec:lift} presents Joukowski spectral lifting, the LiftGCN architecture, and its spectral and computational properties. 
Section~\ref{sec:experiment} evaluates LiftGCN through comparative experiments, depth analysis, and ablation studies. 
Finally, Section~\ref{sec:conclusion} concludes the paper.

\section{Background}\label{sec:background}

Stress fields contain sharp spatial variations that graph smoothing can suppress. We briefly connect their physical regularity to graph-frequency content, then contrast diffusive and unitary propagation.

\subsection{Finite Element Stress Fields and Limited Regularity}

For a linear elastic body occupying $\Omega\subset\mathbb{R}^m$, stress is $\boldsymbol{\sigma}=\mathbb{C}:\boldsymbol{\varepsilon}(\mathbf{u})$, where $\mathbf{u}$ is displacement, $\mathbb{C}$ is the elasticity tensor, and $\boldsymbol{\varepsilon}(\mathbf{u})=(\nabla\mathbf{u}+\nabla\mathbf{u}^{\top})/2$. Equilibrium requires $-\nabla\cdot\boldsymbol{\sigma}=\mathbf{f}$, subject to prescribed displacement and traction conditions. The finite element formulation is based on the weak problem
\begin{equation}
    \int_{\Omega}\boldsymbol{\varepsilon}(\mathbf{v}):\mathbb{C}:\boldsymbol{\varepsilon}(\mathbf{u})\,d\Omega
    =\int_{\Omega}\mathbf{v}\cdot\mathbf{f}\,d\Omega
    +\int_{\Gamma_N}\mathbf{v}\cdot\bar{\mathbf{t}}\,d\Gamma,
    \qquad \forall\mathbf{v}\in V_0,
    \label{eq:weak_equilibrium}
\end{equation}
where $V_0$ contains test displacements that vanish on the prescribed-displacement boundary, and $\bar{\mathbf{t}}$ is the traction on $\Gamma_N$. Under standard assumptions, the energy solution belongs to $H^1(\Omega)$~\citep{bathe1996finite,zienkiewicz2013finite}. Since stress depends on displacement derivatives, its natural regularity is generally $L^2(\Omega)$; global smoothness is not guaranteed.

Holes and smooth fillets can produce large but finite stress gradients, while sharp notches, re-entrant corners, and material interfaces can induce stronger local variations. Near a geometric singularity, the leading stress response may take the form
\begin{equation}
    \boldsymbol{\sigma}(r,\theta)\sim r^{\kappa-1}\boldsymbol{\Phi}(\theta),
    \qquad 0<\kappa<1,
\end{equation}
where $r$ is the distance to the singularity. Such behavior explains why a stress field can be difficult to reconstruct even when the surrounding displacement field appears smooth.

On a mesh graph, let $\mathcal{L}=\mathbf{I}-\widetilde{\mathbf{A}}=\mathbf{Q}\operatorname{diag}(\mu_j)\mathbf{Q}^{\top}$ be the normalized Laplacian. For a nodal stress signal $\mathbf{y}$ with graph Fourier coefficients $\widehat{\mathbf{y}}=\mathbf{Q}^{\top}\mathbf{y}$,
\begin{equation}
    \mathbf{y}^{\top}\mathcal{L}\mathbf{y}
    =\sum_j\mu_j|\widehat y_j|^2.
    \label{eq:dirichlet}
\end{equation}
Higher graph frequencies contribute more strongly to this variation measure. Preserving them helps reconstruct localized stress patterns that a purely smooth representation may miss.

\subsection{Diffusive and Unitary Graph Propagation}

For an undirected graph, let $\widetilde{\mathbf{A}}=\widehat{\mathbf{D}}^{-1/2}\widehat{\mathbf{A}}\widehat{\mathbf{D}}^{-1/2}$ denote a symmetric normalized adjacency, where $\widehat{\mathbf{A}}$ may include self-loops. Its eigenvalues $\lambda_j$ lie in $[-1,1]$. A standard GCN updates features as $\mathbf{H}_{l+1}=\phi(\widetilde{\mathbf{A}}\mathbf{H}_l\mathbf{W}_l)$. Isolating the linear propagation gives
\begin{equation}
    \widetilde{\mathbf{A}}^L=\mathbf{Q}\operatorname{diag}(\lambda_j^L)\mathbf{Q}^{\top},
    \qquad |\lambda_j|^L\to0\quad\text{for }|\lambda_j|<1.
\end{equation}
Repeated aggregation therefore attenuates these modes and favors smoother representations~\citep{li2018deeper,hoang2021revisiting}. In particular, low Laplacian frequencies correspond to adjacency eigenvalues close to one, which decay slowly. Components with smaller eigenvalue magnitudes disappear more quickly. This unequal attenuation can weaken local stress contrasts as the receptive field grows, linking the physical reconstruction problem to the spectral behavior of propagation.

Unitary propagation instead uses
\begin{equation}
    \mathbf{U}(t)=e^{\mathrm{i}t\widetilde{\mathbf{A}}}
    =\mathbf{Q}\operatorname{diag}(e^{\mathrm{i}t\lambda_j})\mathbf{Q}^{\top},
    \qquad |e^{\mathrm{i}t\lambda_j}|=1.
\end{equation}
Each spectral component changes phase without losing magnitude~\citep{UniGCN}. Consequently, $\mathbf{U}(t)^*\mathbf{U}(t)=\mathbf{I}$, so repeated unitary propagation preserves the Euclidean norm of the feature signal. The attraction is that increasing the propagation distance need not progressively erase frequency content. Approximating the matrix exponential requires multiple graph operations, motivating a simpler realization of unit-modulus dynamics.

\section{Joukowski Spectral Lifting and LiftGCN}\label{sec:lift}

LiftGCN obtains stable spectral dynamics through state-space lifting. We derive the real-valued recurrence, combine it with local nonlinear transformations, and establish its energy-preserving structure and propagation cost.

\subsection{Joukowski Spectral Lifting}\label{sec:joukowski}

\begin{figure}[t]
    \centering

    \begin{subfigure}[t]{0.31\textwidth}
        \centering
        \includegraphics[width=\linewidth]
        {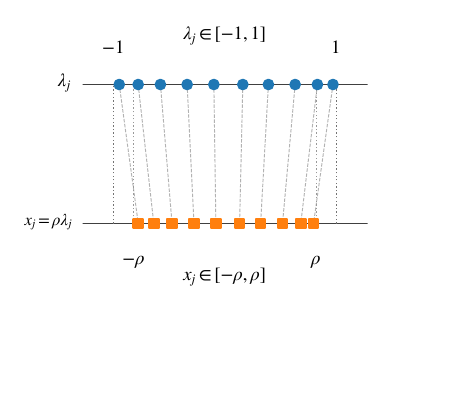}
        \caption{Real graph spectrum.}
        \label{fig:joukowski-spectrum}
    \end{subfigure}
    \hfill
    \begin{subfigure}[t]{0.31\textwidth}
        \centering
        \includegraphics[width=\linewidth]
        {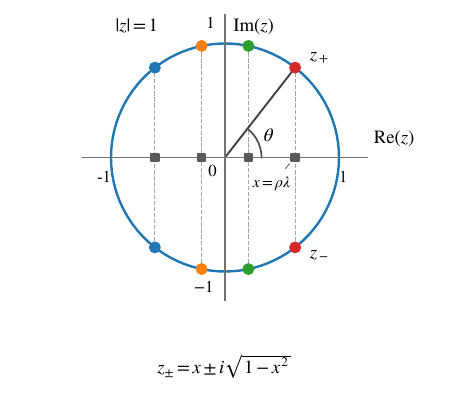}
        \caption{Joukowski spectral lifting.}
        \label{fig:joukowski-circle}
    \end{subfigure}
    \hfill
    \begin{subfigure}[t]{0.31\textwidth}
        \centering
        \includegraphics[width=\linewidth]
        {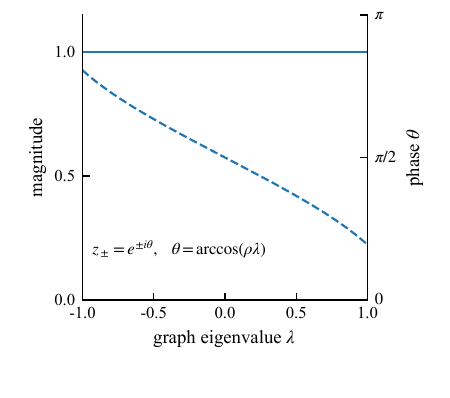}
        \caption{Unit-modulus response.}
        \label{fig:joukowski-response}
    \end{subfigure}

    \caption{Joukowski spectral lifting maps the scaled real graph spectrum to conjugate unit-circle modes, preserving their magnitude while changing their phase.}
    \label{fig:joukowski-lifting}
\end{figure}

The Joukowski map $J(z)=(z+z^{-1})/2$ sends $z=e^{\mathrm{i}\theta}$ to $\cos\theta$~\citep{Joukowski}. Conversely, for $x\in(-1,1)$, solving $z^2-2xz+1=0$ gives
\begin{equation}
    z_{\pm}=x\pm\mathrm{i}\sqrt{1-x^2}=e^{\pm\mathrm{i}\arccos x},
    \qquad |z_{\pm}|=1.
    \label{eq:joukowski_roots}
\end{equation}
Setting $x=\rho_c\lambda_j$ with $|\rho_c|<1$ lifts each graph eigenvalue to two unit-circle modes, as illustrated in Figure~\ref{fig:joukowski-lifting}. Their phases depend on graph frequency and channel, while their magnitudes remain one.

The same quadratic is the characteristic equation of $h_{j,c}^{(l+1)}=2\rho_c\lambda_jh_{j,c}^{(l)}-h_{j,c}^{(l-1)}$. Replacing the eigenvalue by the graph operator yields
\begin{equation}
    \overline{\mathbf{H}}_{l+1}
    =2\widetilde{\mathbf{A}}\mathbf{H}_l\mathbf{R}-\mathbf{H}_{l-1},
    \qquad \mathbf{R}=\operatorname{diag}(\rho_1,\ldots,\rho_d).
    \label{eq:lift_recurrence}
\end{equation}
This recurrence implements the lifted dynamics entirely in real arithmetic. The previous hidden state supplies the second degree of freedom required by the conjugate roots; no complex-valued features or explicit spectral decomposition are needed. We learn $\rho_c=\rho_{\max}\tanh(\theta_c)$ with $0<\rho_{\max}<1$, sharing each coefficient across depth. Thus, channels can learn distinct spectral phases while remaining in the unit-modulus regime. Sharing the coefficients preserves a stationary linear recurrence, while channel-wise adaptation lets different features use different frequency-dependent phases.

\subsection{LiftGCN Architecture}\label{sec:lift_architecture}

\begin{figure}[t]
  \centering
  \includegraphics[width=\textwidth]
  {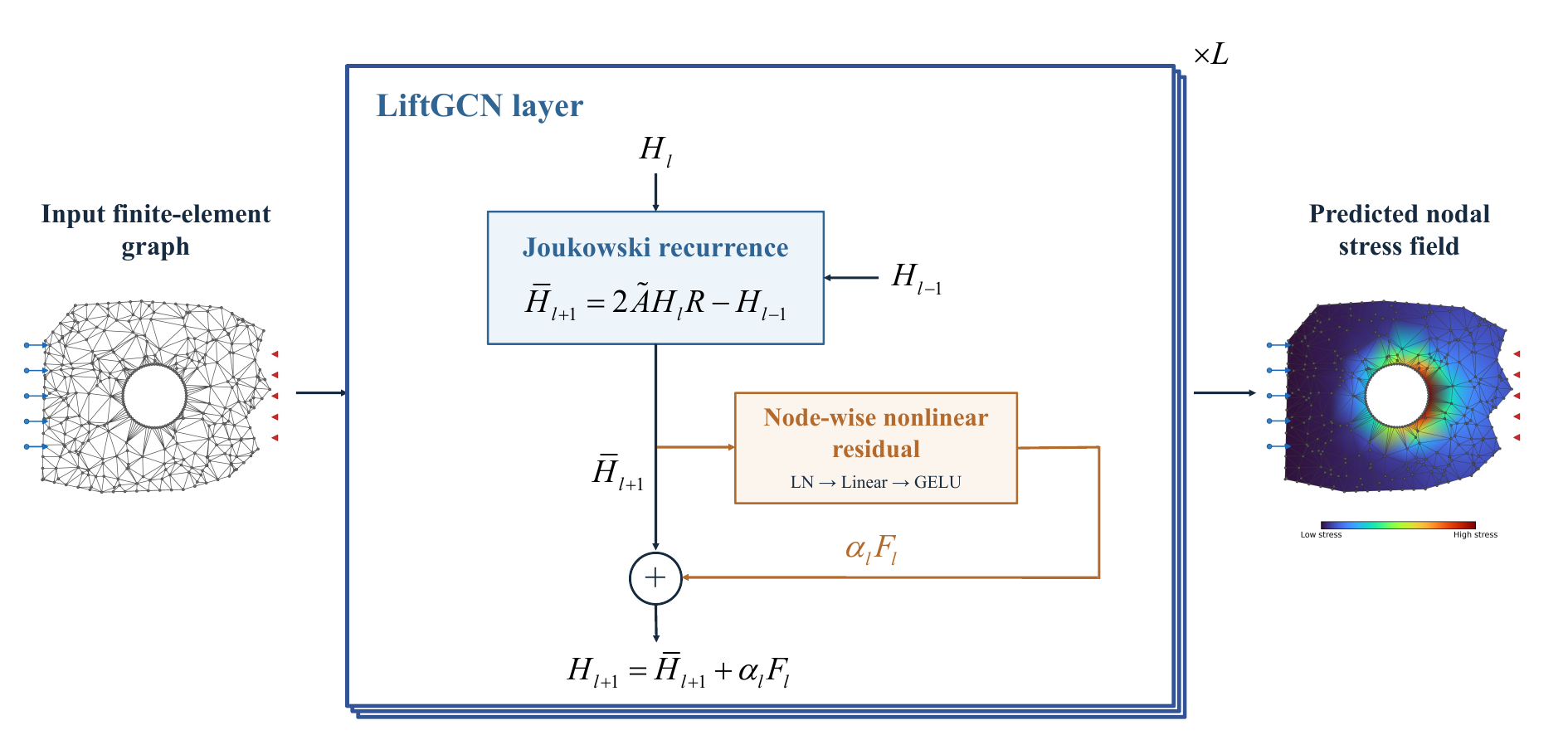}
  \caption{LiftGCN combines second-order Joukowski propagation with node-wise nonlinear residuals. Each layer uses one sparse neighborhood aggregation and the previous hidden state.}
  \label{fig:LiftGCN_pipeline}
\end{figure}

An input encoder maps node features to $\mathbf{H}_0\in\mathbb{R}^{n\times d}$. The first propagation is $\overline{\mathbf{H}}_1=\widetilde{\mathbf{A}}\mathbf{H}_0\mathbf{R}$; subsequent steps use Equation~\ref{eq:lift_recurrence}. Each step then applies a node-wise residual update,
\begin{equation}
    \mathbf{H}_{l+1}=\overline{\mathbf{H}}_{l+1}
    +\alpha_l\operatorname{GELU}\!\left(\operatorname{LN}(\overline{\mathbf{H}}_{l+1})\mathbf{W}_l+\mathbf{b}_l\right),
    \label{eq:lift_layer}
\end{equation}
where $\mathbf{W}_l$ and $\mathbf{b}_l$ mix channels, and $\alpha_l$ is a learnable residual scale. A linear readout maps the final two-state representation $[\mathbf{H}_L,\mathbf{H}_{L-1}]$ to nodal stress. Figure~\ref{fig:LiftGCN_pipeline} summarizes the propagation and residual blocks.

The recurrence controls inter-node information transport, while the residuals learn local nonlinear corrections. Normalization and channel mixing act independently at each node, and the learned residual scale controls the strength of each correction. The two-state readout exposes both states of the recurrence to the predictor. This separation adds expressive feature transformations without introducing extra graph aggregations.

\subsection{Spectral and Computational Properties}\label{sec:lift_properties}

For each graph mode and channel, the linear recurrence has roots $e^{\pm\mathrm{i}\arccos(\rho_c\lambda_j)}$. Their unit magnitude avoids the exponential decay associated with repeated adjacency multiplication. The recurrence also preserves a quadratic energy in its augmented state. For a single channel with coefficient $\rho$, write $\mathbf{Z}_l=[\mathbf{h}_l^{\top},\mathbf{h}_{l-1}^{\top}]^{\top}$ and $\mathbf{Z}_{l+1}=\mathbf{U}_{\rho}\mathbf{Z}_l$, where
\begin{equation}
    \mathbf{U}_{\rho}=\begin{bmatrix}2\rho\widetilde{\mathbf{A}}&-\mathbf{I}\\\mathbf{I}&\mathbf{0}\end{bmatrix},
    \qquad
    \mathbf{G}_{\rho}=\begin{bmatrix}\mathbf{I}&-\rho\widetilde{\mathbf{A}}\\-\rho\widetilde{\mathbf{A}}&\mathbf{I}\end{bmatrix}.
\end{equation}
The eigenvalues of $\mathbf{G}_{\rho}$ are $1\pm\rho\lambda_j>0$, and direct multiplication gives
\begin{equation}
    \mathbf{U}_{\rho}^{\top}\mathbf{G}_{\rho}\mathbf{U}_{\rho}=\mathbf{G}_{\rho},
    \qquad \|\mathbf{Z}_{l+1}\|_{\mathbf{G}_{\rho}}^2=\|\mathbf{Z}_l\|_{\mathbf{G}_{\rho}}^2.
    \label{eq:energy_preservation}
\end{equation}
The conserved quantity can also be written directly in terms of consecutive hidden states:
\begin{equation}
    \|\mathbf{Z}_l\|_{\mathbf{G}_{\rho}}^2
    =\|\mathbf{h}_l\|_2^2+\|\mathbf{h}_{l-1}\|_2^2
    -2\rho\mathbf{h}_l^{\top}\widetilde{\mathbf{A}}\mathbf{h}_{l-1}.
\end{equation}
Energy is therefore preserved in the pair of states, allowing information to move between them as propagation proceeds. This result applies independently to every channel of the linear backbone. It provides a non-dissipative basis for propagation, to which the learned residuals add nonlinear corrections.

Each layer uses one sparse multiplication by $\widetilde{\mathbf{A}}$, giving $O(ed)$ graph propagation cost rather than the $O(Ked)$ cost of a $K$-step matrix-function approximation. Node-wise channel mixing costs $O(nd^2)$ in both cases. LiftGCN therefore reduces the graph propagation work while retaining one-hop communication and locally stored second-order memory.

This locality is useful on large meshes: each node exchanges only its current representation with immediate neighbors and stores its own previous state. Stable spectral dynamics are obtained by augmenting the state, without expanding the communication neighborhood within a layer. Additional layers can then extend the receptive field through the same sparse connectivity.

\raggedbottom
\section{Experiment}\label{sec:experiment}

We evaluate LiftGCN in terms of whole-field accuracy, stress-concentration reconstruction, and inference efficiency. All experiments are conducted on a workstation equipped with a single NVIDIA RTX PRO 6000 GPU. All prediction metrics reported in the tables and quantitative figures are averaged over 10 repeated experiments. We first introduce the dataset and evaluation metrics, then compare LiftGCN with representative baselines and analyze its sensitivity to propagation depth, followed by ablation studies on Joukowski propagation, spectral scaling, and local feature transformations. Appendix~\ref{app:custom_results} extends the comparisons to two additional geometries, with data and training details in Appendices~\ref{app:datasets} and~\ref{app:settings}.

\subsection{Dataset and Evaluation Metrics}\label{sec:experimental_setup}

We use \textit{Mines\_Paris\_Biaxial\_Specimen}, comprising 100 finite element simulations of a biaxial specimen with random elastic properties~\citep{kerfriden2022biaxial}. The task is to predict nodal von Mises stress from geometry, material properties, and mesh connectivity. Appendix~\ref{app:mines} specifies the inputs and mesh statistics, and Appendix~\ref{app:graph_construction} describes graph construction. Training settings and checkpoint selection based on test-set loss are detailed in Appendix~\ref{app:settings}.

We measure whole-field accuracy with SNR and $R^2$. For reference stresses $y_i$ and predictions $\hat y_i$, pooling evaluation nodes within each repetition gives
\begin{equation}
    \mathrm{SNR}=10\log_{10}\frac{\sum_i y_i^2}{\sum_i(\hat y_i-y_i)^2},
    \qquad R^2=1-\frac{\sum_i(\hat y_i-y_i)^2}{\sum_i(y_i-\bar y)^2},
    \label{eq:global_metrics}
\end{equation}
where $\bar y$ is the pooled reference mean. To evaluate concentrations, let $S_k$ contain the top $k\%$ of nodes by reference stress magnitude on each mesh, with $k\in\{1,5,10\}$. Peak-region accuracy is
\begin{equation}
    \mathrm{PSNR}@k\%=10\log_{10}\frac{\max_{i\in S_k}|y_i|^2+\epsilon}
    {|S_k|^{-1}\sum_{i\in S_k}(\hat y_i-y_i)^2+\epsilon}.
    \label{eq:peak_metric}
\end{equation}
For undirected edges $E_k$ touching $S_k$, define $g_{ij}(y)=(y_i-y_j)/\max(\|\mathbf{x}_i-\mathbf{x}_j\|_2,\epsilon)$ using node coordinates $\mathbf{x}_i$. High-stress gradient error is
\begin{equation}
    \mathrm{HSG\text{-}NMSE}@k\%=
    \frac{\sum_{(i,j)\in E_k}[g_{ij}(\hat y)-g_{ij}(y)]^2}
    {\max(\sum_{(i,j)\in E_k}g_{ij}(y)^2,\epsilon)}.
    \label{eq:hsg_metric}
\end{equation}
Here $\epsilon$ ensures numerical stability. PSNR and HSG-NMSE are averaged over evaluation meshes before averaging across the 10 repetitions. Higher SNR, $R^2$, and PSNR and lower HSG-NMSE are better. Together, these metrics assess the overall stress field, its peaks, and the sharp variations around them. We also report parameter counts and Latency, defined as the average forward inference time per sample (one complete mesh), measured separately from the repeated training experiments.

\subsection{Main Comparisons, Efficiency, and Depth Sensitivity}\label{sec:main_comparisons}

Table~\ref{tab:main_results} compares LiftGCN with conventional message-passing, mesh-based, dynamical, spectral, and attention-based graph models. At comparable parameter counts, small LiftGCN achieves the lowest latency and improves both whole-field and local accuracy over GCN. It also matches or slightly improves UniGCN's peak-region PSNR with much faster inference. These 10-run mean results show that the lifted recurrence can retain useful local stress information at low propagation cost.

\begin{table}[t]
    \centering
    \caption{Main comparison. Prediction metrics are means over 10 repeated experiments. 
    \textbf{Bold} marks the best prediction metric or lowest latency. 
    Hybrid LG denotes Hybrid Local--Global Attention GNN, and Adv-GCN denotes GCN with adversarial training. 
    Parentheses indicate the number of discriminator parameters.}
    \label{tab:main_results}
    \begingroup
    \footnotesize
    \setlength{\tabcolsep}{3.5pt}
    \renewcommand{\arraystretch}{1.1}

    \textbf{(a) Whole-field accuracy and computational cost}\\[3pt]
    \begin{tabular*}{\textwidth}{@{\extracolsep{\fill}}lrcrrr@{}}
        \toprule
        Model 
        & Parameters 
        & \begin{tabular}[c]{@{}c@{}}Graph propagation\\complexity\end{tabular}
        & \begin{tabular}[c]{@{}c@{}}Latency$\downarrow$\\(ms)\end{tabular}
        & \begin{tabular}[c]{@{}c@{}}SNR$\uparrow$\\(dB)\end{tabular}
        & $R^2\uparrow$ \\
        \midrule
        GCN~\citep{GCN} 
        & 444k & $\mathcal{O}(ed)$ 
        & 1.719 & 15.685 & 0.85744 \\

        Adv-GCN~\citep{AdvGCN} 
        & {444k (+67k)} & $\mathcal{O}(ed)$ 
        & 1.719 & 15.605 & 0.85479 \\

        MeshGraphNets~\citep{MeshGraphNets} 
        & 401k & $\mathcal{O}(ed^2)$ 
        & 6.173 & 16.008 & 0.86764 \\

        GCNII~\citep{GCNII} 
        & 439k & $\mathcal{O}(ed)$ 
        & 1.639 & 15.095 & 0.83670 \\

        GraphCON~\citep{GraphCON} 
        & 439k & $\mathcal{O}(ed)$ 
        & 1.703 & 13.307 & 0.75349 \\

        CayleyNet~\citep{Cayleynet} 
        & 428k & $\mathcal{O}(rJed)$ 
        & 28.164 & 16.126 & 0.87109 \\

        EWGNN~\citep{Wave-driven-GNN} 
        & 462k & $\mathcal{O}(n^2d)$ 
        & 4.393 & 15.212 & 0.84098 \\

        GUMP~\citep{GUMP} 
        & 439k & $\mathcal{O}(M_Ld)^{\dagger}$ 
        & 48.527 & 13.441 & 0.76017 \\

        Hybrid LG~\citep{HybridLG} 
        & 438k & $\mathcal{O}(ed^2+n^2d/M)$ 
        & 13.365 & \textbf{17.104} & \textbf{0.89607} \\

        A-DGN~\citep{A-DGN} 
        & 403k & $\mathcal{O}(ed)$ 
        & 2.211 & 15.671 & 0.85696 \\

        UniGCN~\citep{UniGCN} 
        & 446k & $\mathcal{O}(Ked)$ 
        & 15.000 & 16.104 & 0.87053 \\
        \midrule

        LiftGCN (ours), small 
        & 443k & $\mathcal{O}(ed)$ 
        & \textbf{1.528} & 15.919 & 0.86487 \\

        LiftGCN (ours), medium 
        & 3299k & $\mathcal{O}(ed)$ 
        & 5.041 & 16.610 & 0.88475 \\

        LiftGCN (ours), large 
        & 16.4M & $\mathcal{O}(ed)$ 
        & 13.851 & 16.821 & 0.89022 \\
        \bottomrule
    \end{tabular*}

    \vspace{7pt}
    \textbf{(b) Stress-concentration reconstruction}\\[3pt]
    \begin{tabular*}{\textwidth}{@{\extracolsep{\fill}}lrrrrrr@{}}
        \toprule
        \multirow{2}{*}{Model}
        & \multicolumn{3}{c}{PSNR (dB)$\uparrow$}
        & \multicolumn{3}{c}{HSG-NMSE$\downarrow$} \\
        \cmidrule(lr){2-4}\cmidrule(l){5-7}
        & @1\% & @5\% & @10\% & @1\% & @5\% & @10\% \\
        \midrule
        GCN~\citep{GCN} 
        & 18.886 & 21.779 & 23.406 
        & 0.145440 & 0.117880 & 0.101360 \\

        Adv-GCN~\citep{AdvGCN} 
        & 19.094 & 21.947 & 23.417 
        & 0.142850 & 0.117070 & 0.104030 \\

        MeshGraphNets~\citep{MeshGraphNets} 
        & 19.737 & 22.282 & 23.772 
        & 0.089125 & 0.082616 & 0.076469 \\

        GCNII~\citep{GCNII} 
        & 18.662 & 21.225 & 22.710 
        & 0.196790 & 0.163430 & 0.148320 \\

        GraphCON~\citep{GraphCON} 
        & 17.062 & 19.514 & 20.868 
        & 0.415490 & 0.403890 & 0.397550 \\

        CayleyNet~\citep{Cayleynet} 
        & 20.783 & 22.281 & 23.589 
        & 0.083929 & 0.098303 & 0.098552 \\

        EWGNN~\citep{Wave-driven-GNN} 
        & 18.746 & 21.312 & 22.943 
        & 0.150180 & 0.132090 & 0.117000 \\

        GUMP~\citep{GUMP} 
        & 17.478 & 19.888 & 21.232 
        & 0.350990 & 0.379950 & 0.388420 \\

        Hybrid LG~\citep{HybridLG} 
        & 21.501 & 23.813 & \textbf{25.138} 
        & 0.072304 & 0.071222 & 0.067561 \\

        A-DGN~\citep{A-DGN} 
        & 19.086 & 21.778 & 23.338 
        & 0.095384 & 0.090815 & 0.083688 \\

        UniGCN~\citep{UniGCN} 
        & 20.096 & 22.526 & 23.974 
        & 0.073533 & 0.072321 & 0.067939 \\
        \midrule

        LiftGCN (ours), small 
        & 20.327 & 22.615 & 23.975 
        & 0.087084 & 0.081528 & 0.075953 \\

        LiftGCN (ours), medium 
        & 21.884 & 23.744 & 24.871 
        & 0.058066 & 0.061305 & 0.060404 \\

        LiftGCN (ours), large 
        & \textbf{22.201} & \textbf{24.035} & 25.086 
        & \textbf{0.053120} & \textbf{0.057062} & \textbf{0.057472} \\
        \bottomrule
    \end{tabular*}

    \vspace{3pt}
    \begin{minipage}{\textwidth}
        \scriptsize
        Propagation complexity is reported per layer and excludes purely node-wise feature
        transformations. Here $n$, $e$, and $d$ denote the numbers of nodes, edges, and hidden
        features, respectively. $K$ is the Taylor truncation order in UniGCN; $r$ and $J$ are
        the Cayley filter order and number of Jacobi iterations in CayleyNet; and $M$ is the
        global-attention interval in Hybrid LG, whose cost is shown as the amortized per-layer
        complexity. For GUMP, $M_L=\sum_{v\in V}\deg(v)^2$ is the number of admissible
        transitions in the directed line graph. $^\dagger$GUMP additionally requires unitary
        operator construction, whose $K_G$-step blockwise Newton--Schulz projection costs
        $\mathcal{O}\!\left(K_G\sum_{v\in V}\deg(v)^3\right)$.
    \end{minipage}

    \endgroup
\end{table}

Increasing model capacity makes further use of this efficiency. Medium LiftGCN outperforms MeshGraphNets and UniGCN across the reported prediction metrics while remaining faster. With a latency budget comparable to Hybrid Local--Global Attention GNN and UniGCN, large LiftGCN leads the peak-region PSNR at the two smallest thresholds and all high-stress gradient metrics. Its strongest gains lie in the localized structures that motivate the model, alongside competitive overall accuracy.

\begin{figure}[t]
    \centering
    \includegraphics[width=\textwidth]{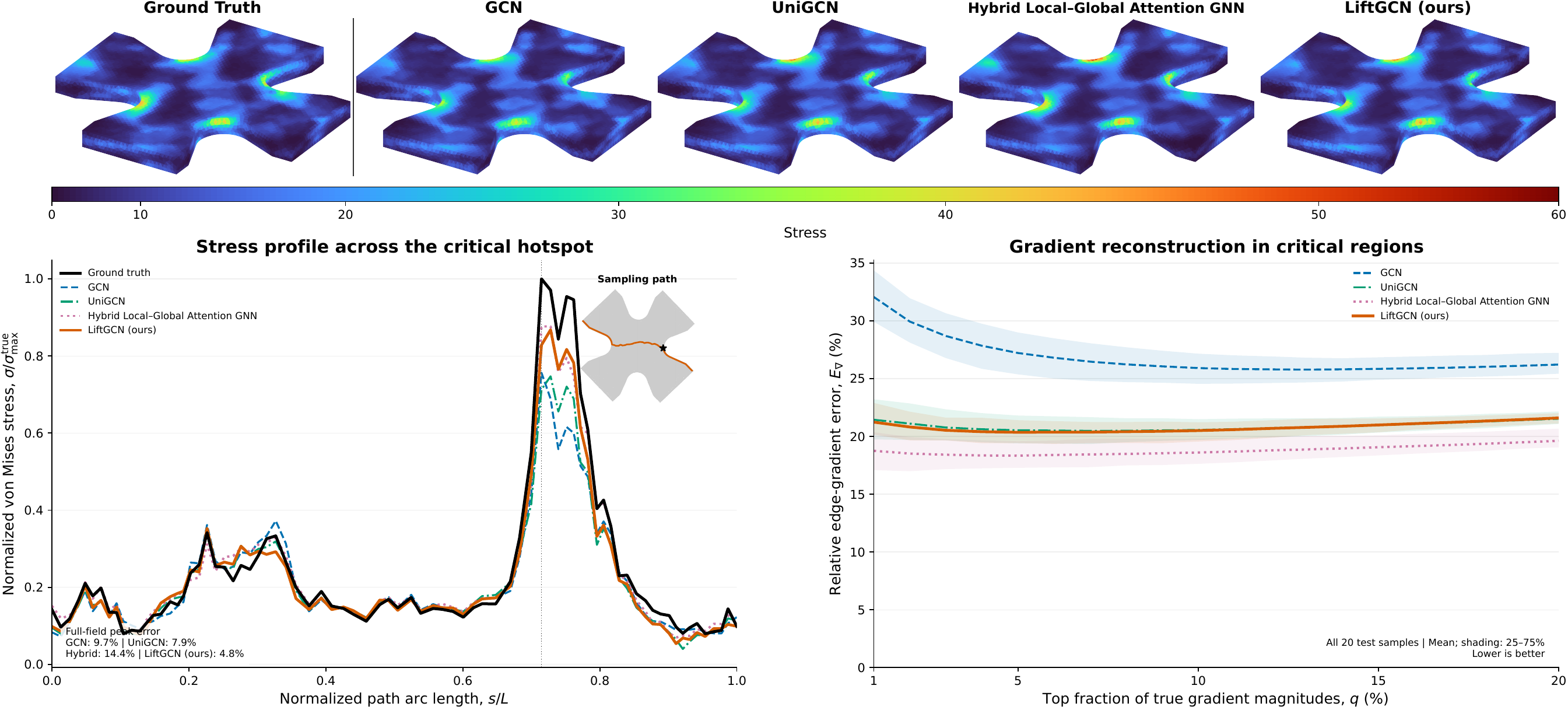}
    \caption{Stress-field reconstruction and hotspot analysis: representative stress fields, a hotspot stress profile, and gradient errors across the test set. LiftGCN preserves localized concentrations and spatial variations.}
    \label{fig:stress_comparison}
\end{figure}

Figure~\ref{fig:stress_comparison} makes this advantage visible. LiftGCN reproduces compact high-stress bands near curved boundaries and preserves heterogeneous patterns within the specimen. These local details agree with the peak and gradient improvements observed in the 10-run mean metrics.

\begin{figure}[t]
    \centering
    \begin{subfigure}[t]{0.49\textwidth}
        \centering
        \includegraphics[width=\linewidth]{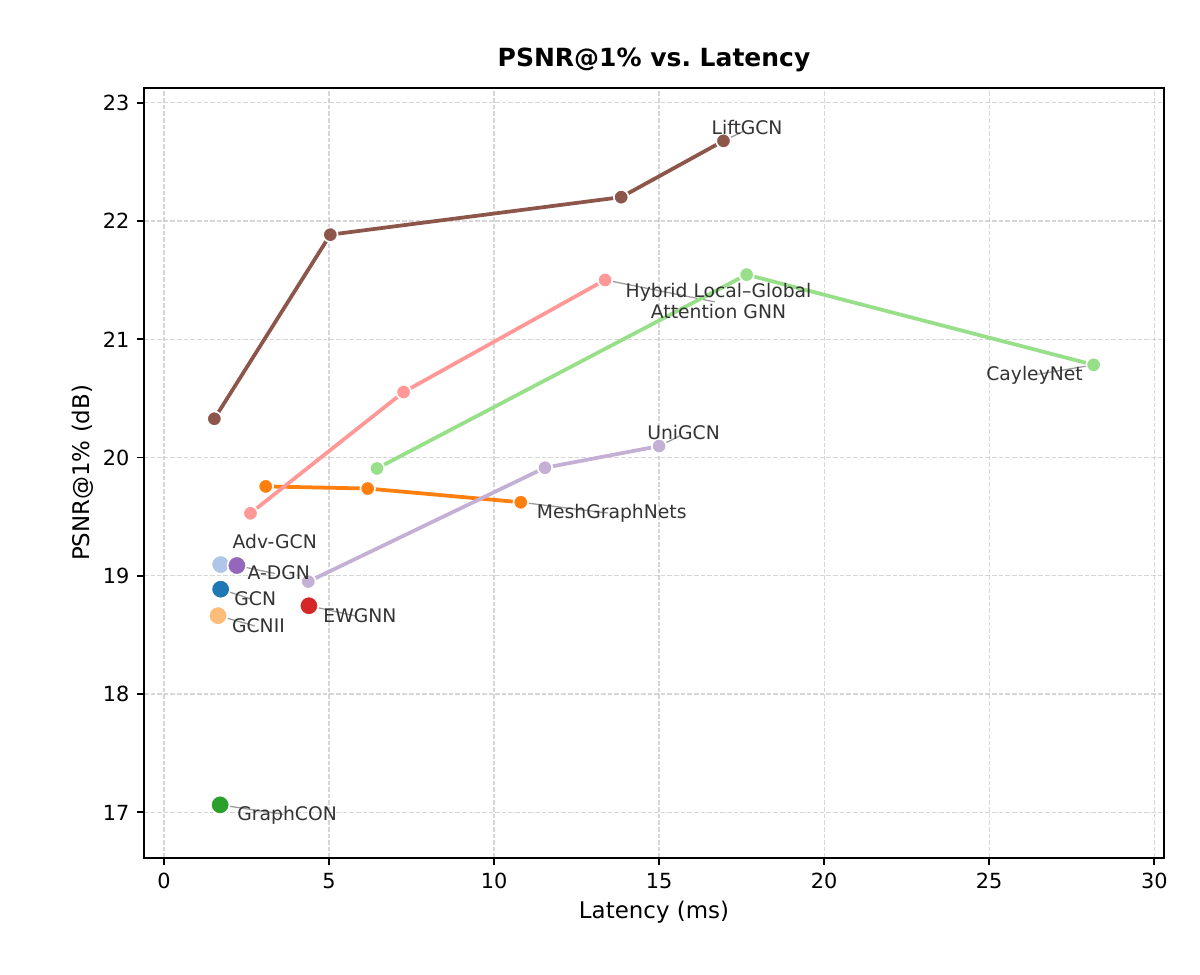}
        \caption{Peak-stress accuracy versus latency.}
    \end{subfigure}\hfill
    \begin{subfigure}[t]{0.49\textwidth}
        \centering
        \includegraphics[width=\linewidth]{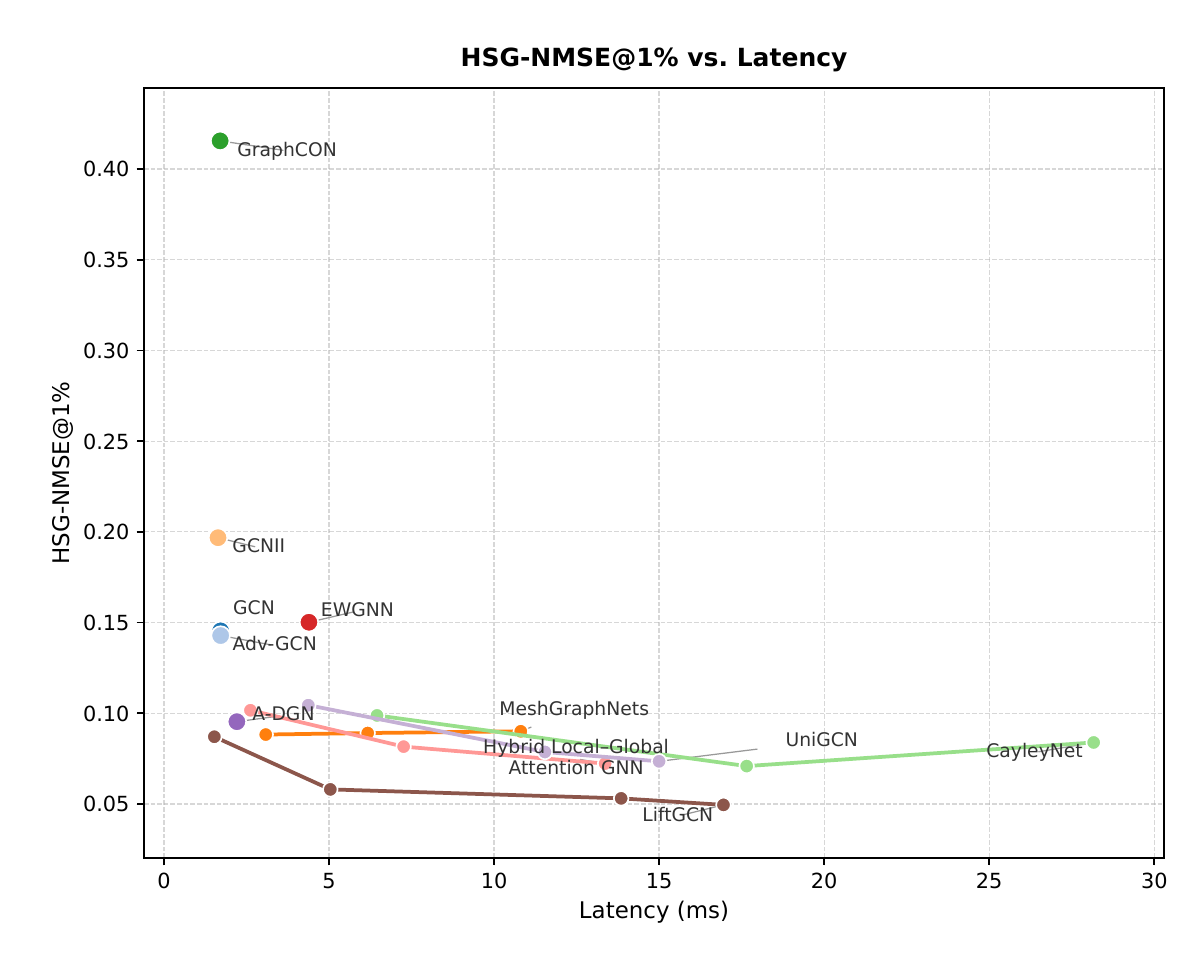}
        \caption{High-stress gradient error versus latency.}
    \end{subfigure}
    \caption{Accuracy--latency trade-offs. Prediction metrics are averaged over 10 repeated experiments. LiftGCN (ours) achieves strong peak-stress and local-gradient reconstruction across inference budgets.}
    \label{fig:accuracy_latency}
\end{figure}

The accuracy--latency curves in Figure~\ref{fig:accuracy_latency} show a favorable trade-off across model sizes. LiftGCN already reconstructs stress concentrations well at low latency, and additional capacity steadily improves peak and gradient accuracy. The benefit of cheaper propagation is therefore practical: more of the inference budget can support feature learning. Comparisons on the connecting-lug and elbow-bracket datasets in Appendix~\ref{app:custom_results} further assess this trade-off across geometries.

The two local metrics reveal complementary aspects of this trade-off. Higher peak-region PSNR indicates closer agreement with the stress values in highly loaded regions, while lower HSG-NMSE indicates better recovery of how stress changes around them. LiftGCN improves both as capacity increases. Its advantage therefore extends beyond matching peak values to recovering the surrounding spatial structure, which is central to a faithful stress-field surrogate.

\begin{figure}[t]
    \centering
    \begin{subfigure}[t]{0.32\textwidth}
        \centering
        \includegraphics[width=\linewidth]{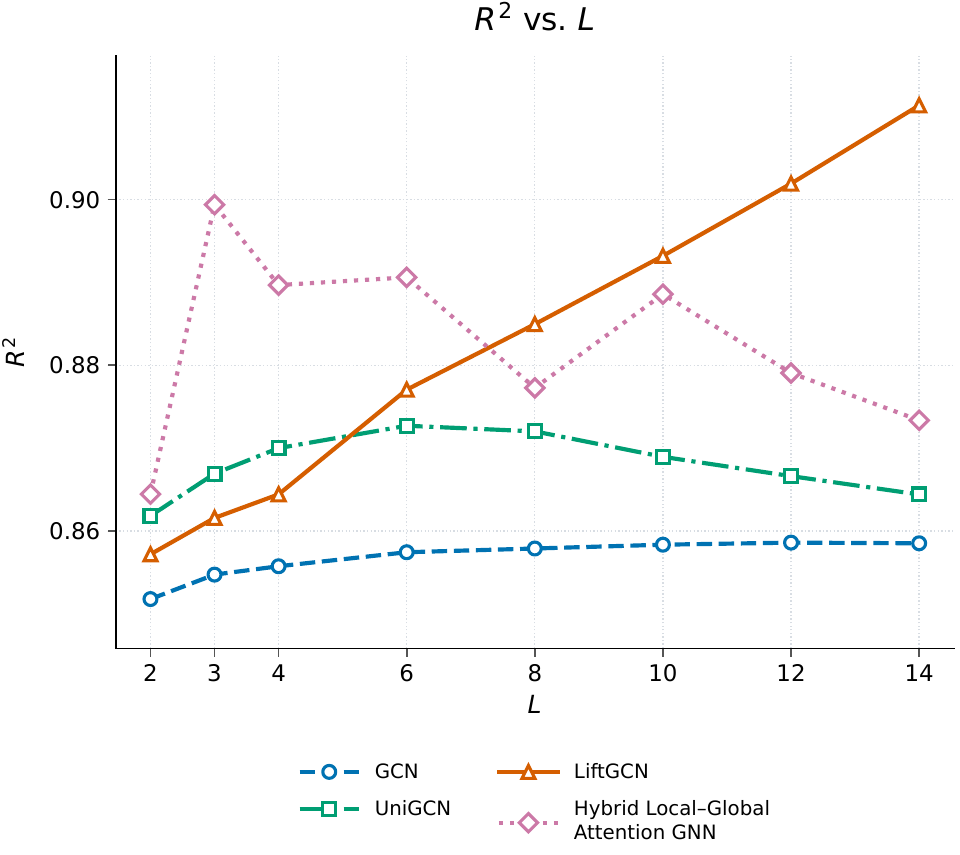}
        \caption{Whole-field $R^2$.}
    \end{subfigure}\hfill
    \begin{subfigure}[t]{0.32\textwidth}
        \centering
        \includegraphics[width=\linewidth]{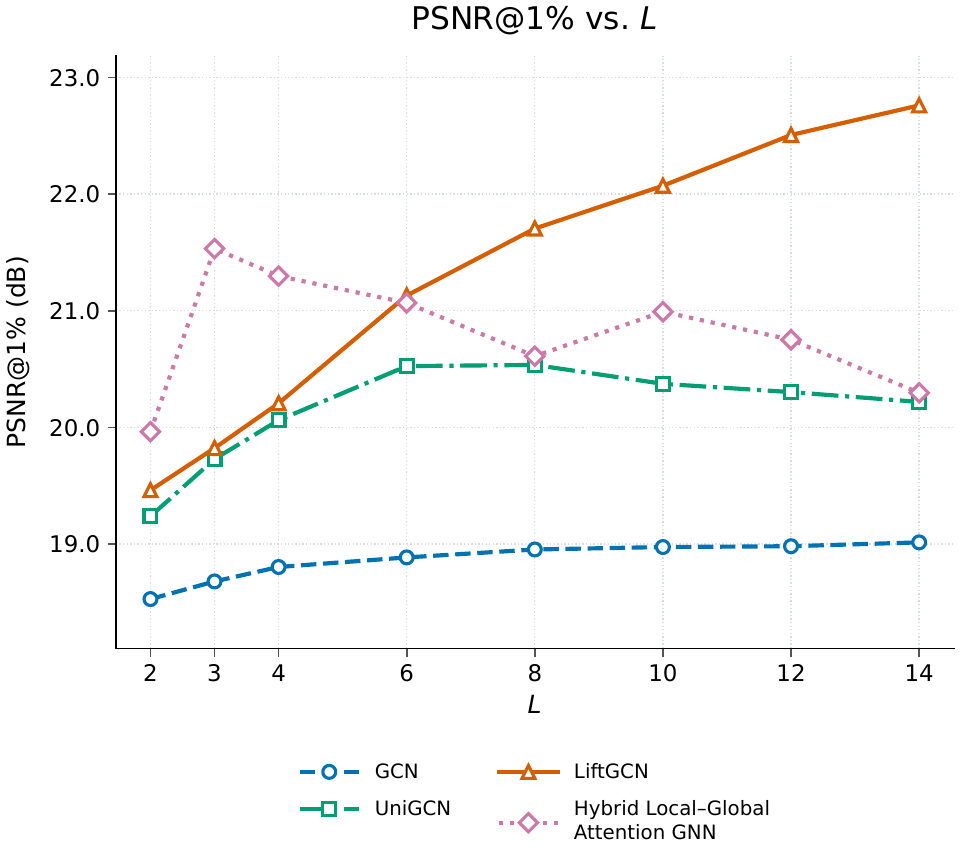}
        \caption{PSNR@1\%.}
    \end{subfigure}\hfill
    \begin{subfigure}[t]{0.32\textwidth}
        \centering
        \includegraphics[width=\linewidth]{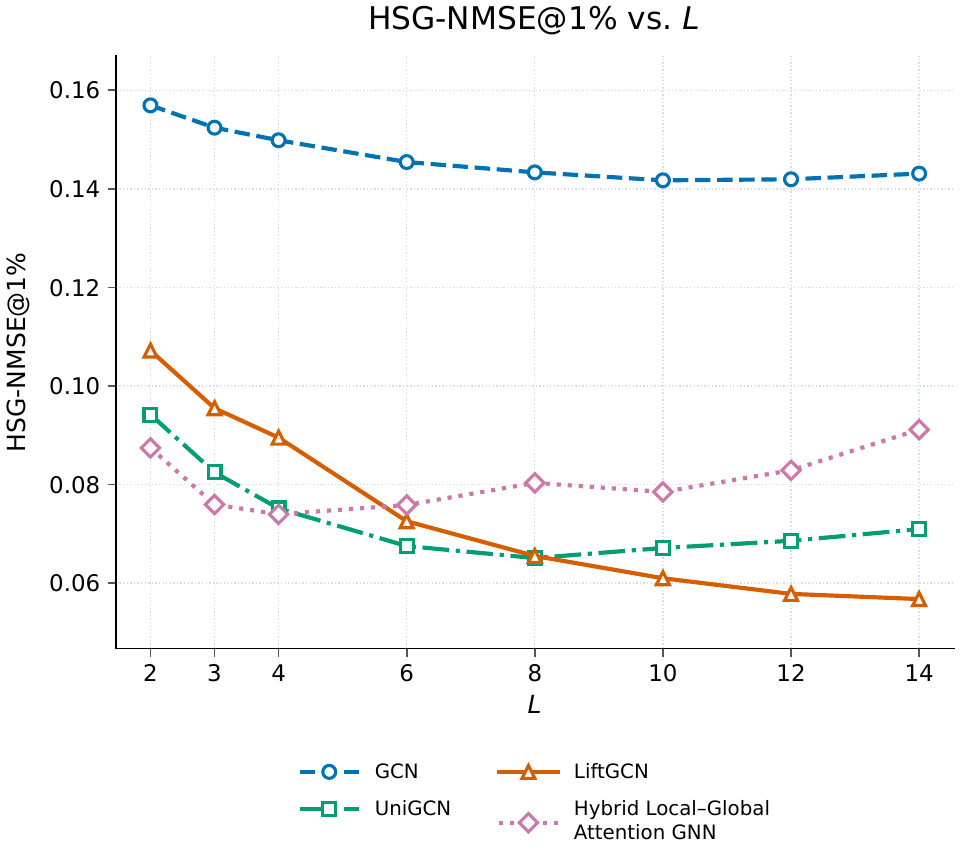}
        \caption{HSG-NMSE@1\%.}
    \end{subfigure}
    \caption{Depth sensitivity. Every point is a mean over 10 repeated experiments. LiftGCN (ours) improves both whole-field and local reconstruction as propagation depth increases.}
    \label{fig:depth_sensitivity}
\end{figure}

Figure~\ref{fig:depth_sensitivity} shows that LiftGCN benefits consistently from greater depth across the tested range. Whole-field and peak accuracy improve while local gradient error decreases. GCN largely saturates, and the unitary and attention baselines perform best at shallower or intermediate depths. These 10-run averages suggest that LiftGCN can incorporate wider spatial context while retaining the local variations needed for stress reconstruction.

This joint improvement matters because increasing the receptive field is useful only if the additional context remains informative. The depth curves show that LiftGCN can improve the broad stress distribution without sacrificing the sharp features measured by the local metrics. This behavior is consistent with the role of the lifted backbone: it supports repeated information transport while the residual blocks progressively adapt the features to the prediction task.

\subsection{Ablation Studies}\label{sec:ablation}

We test four variants, retaining the input encoder, depth, and two-state readout throughout. \textit{w/o Joukowski} replaces the lifted recurrence with first-order aggregation, while \textit{Fixed-$\rho$} fixes the spectral coefficients and retains the local residual blocks. \textit{w/o nonlinear} removes only GELU from each propagation-layer residual block, retaining LayerNorm, channel mixing, and the residual connection. \textit{Pure Joukowski} removes these entire blocks, leaving the hidden states to follow the second-order recurrence. Both variants retain learnable channel-wise spectral coefficients and the nonlinear input encoder. Table~\ref{tab:ablation} reports the means over 10 repetitions.

\begin{table}[t]
    \centering
    \caption{Ablation results averaged over 10 repeated experiments. Bold marks the best mean. Accurate stress reconstruction benefits from both Joukowski propagation and local nonlinear transformations; learnable spectral scaling provides further refinement.}
    \label{tab:ablation}
    \small
    \setlength{\tabcolsep}{4pt}
    \begin{tabular*}{\textwidth}{@{\extracolsep{\fill}}lrrrr@{}}
        \toprule
        Model & SNR (dB)$\uparrow$ & $R^2\uparrow$ & PSNR@1\% (dB)$\uparrow$ & HSG-NMSE@1\%$\downarrow$ \\
        \midrule
        LiftGCN (ours) & \textbf{16.329} & \textbf{0.87705} & \textbf{21.131} & \textbf{0.072548} \\
        w/o Joukowski & 14.227 & 0.80057 & 18.027 & 0.275960 \\
        Fixed-$\rho$ ($0.9$) & 16.327 & 0.87698 & 21.098 & 0.073043 \\
        w/o nonlinear & 15.342 & 0.84570 & 19.387 & 0.120570 \\
        Pure Joukowski & 12.897 & 0.72913 & 12.050 & 0.407700 \\
        \bottomrule
    \end{tabular*}
\end{table}

Removing Joukowski propagation reduces accuracy across all metrics, especially in local gradient reconstruction, even with the local transformations retained. The lifted recurrence therefore plays an important role in preserving stress structure. Fixed spectral scaling retains most of the full model's performance, with learnable coefficients providing a small additional improvement.

Removing only GELU also weakens whole-field and local reconstruction, showing that nonlinear activation improves how propagated features are used. GELU has no trainable parameters, so this comparison preserves model size and isolates the activation within the local residual blocks. Removing the entire local residual module causes the largest degradation, particularly in peak-stress and gradient accuracy. The much stronger results with channel mixing and residual updates retained show the value of feature transformations between propagation steps. Together, the 10-run means support the combined design: Joukowski propagation preserves information, while local nonlinear transformations turn it into accurate stress predictions.

\section{Conclusion}\label{sec:conclusion}

LiftGCN realizes Joukowski spectral lifting through a simple real-valued second-order recurrence. Its linear backbone preserves a quadratic energy, and each layer requires only one sparse neighborhood aggregation. Experiments averaged over 10 repetitions show that this design combines efficient inference with accurate reconstruction of stress concentrations and local gradients. The model uses additional capacity and depth effectively, while ablations demonstrate the complementary benefits of lifted propagation and local nonlinear transformations. These findings support spectral lifting as an efficient basis for graph-based stress prediction.

More broadly, the method connects information-preserving dynamics with the sparse communication structure of finite element meshes. It offers a practical route to retaining local physical detail as graph models aggregate wider context. Extending this approach to broader geometries, material models, and physical prediction tasks is a natural direction for future work.

\clearpage
\subsection*{AI use statement}
Generative AI tools, including ChatGPT, were used to assist with the refinement of theoretical formulations and mathematical presentation, experimental design, software implementation, and interpretation of experimental results. They were also used to assist with literature search and summarization, drafting and editing parts of the manuscript, improving readability, and preparing research code and scientific figures. All AI-assisted mathematical derivations, claims, references, code, and written content were independently reviewed, verified, and revised by the authors. The generated and modified code was tested by the authors, and all reported experimental results were obtained from experiments conducted and evaluated by the authors rather than generated by AI. The authors take full responsibility for the methodology, results, claims, and final content of this work.

\subsection*{Reproducibility statement}

Section~\ref{sec:lift} specifies the LiftGCN recurrence, architecture, and assumptions for the energy-preservation result. Appendix~\ref{app:datasets} documents dataset sizes, simulation parameters, stress extraction, and graph construction. Appendix~\ref{app:custom_results} provides the additional benchmark results, and Appendix~\ref{app:settings} records normalization, optimization, repeated splits, checkpoint selection, and inference timing. The public Mines Paris data are identified by their dataset citation~\citep{kerfriden2022biaxial}. Our codee is available at \url{https://github.com/ChenZeng001/LiftGCN}.

\bibliography{references}
\bibliographystyle{iclr2027_conference}

\clearpage
\appendix
\section{Datasets and Graph Representation}\label{app:datasets}

This appendix specifies the finite element data used in Section~\ref{sec:experimental_setup}, including the quantities actually supplied to the networks. Table~\ref{tab:dataset_statistics} summarizes all complete samples in the processed datasets. Counts are measured from the coordinate and element-connectivity files; graph edges follow the constructions in Appendix~\ref{app:graph_construction}. An edge is counted once as an undirected pair, before adding self-loops or storing both message-passing directions.

\begin{table}[htbp]
\centering
\caption{Dataset statistics. Variable mesh sizes are reported as minimum--maximum, with arithmetic means on separate rows. The feature dimension includes the degree feature.}
\label{tab:dataset_statistics}
\small
\setlength{\tabcolsep}{4pt}
\begin{tabular*}{\textwidth}{@{\extracolsep{\fill}}lrrr@{}}
\toprule
Property & Mines Paris & Connecting lug & Elbow bracket \\
\midrule
Samples & 100 & 225 & 288 \\
Nodes per mesh & 10,502 & 9,344--11,762 & 10,282--17,944 \\
Mean nodes & 10,502 & 10,683.72 & 13,878.89 \\
Elements per mesh & 43,399 & 5,778--7,515 & 6,088--10,936 \\
Mean elements & 43,399 & 6,735.65 & 8,354.37 \\
Undirected graph edges & 60,544 & 15,918--20,132 & 17,396--30,500 \\
Mean graph edges & 60,544 & 18,251.52 & 23,544.44 \\
Element nodes & 4 & 10 (C3D10M) & 10 (C3D10M) \\
Input/output channels & $5/1$ & $4/1$ & $4/1$ \\
Shared mesh across samples & Yes & No & No \\
Training/evaluation samples & $80/20$ & $180/45$ & $230/58$ \\
\bottomrule
\end{tabular*}
\end{table}

\subsection{Mines Paris Biaxial Specimen}\label{app:mines}

We use the 100 random-material realizations of the public Mines Paris biaxial-specimen dataset, version V0.9~\citep{kerfriden2022biaxial}. The processed samples have identifiers 1--100; the homogeneous reference realization indexed by 0 in the original archive is not included. All 100 processed coordinate files are identical, as are all 100 connectivity files. The common three-dimensional mesh has 10,502 nodes and 43,399 four-node tetrahedra. Its coordinate bounds are $[-1.5,1.5]\times[-1.5,1.5]\times[0,0.2]$ in the source coordinate units. The original VTK files and the processed CSV files agree on the mesh size. Each realization changes the spatial Young's-modulus field on this fixed specimen. The nodal modulus values across the processed realizations span approximately $[2.9708,129.9183]$ in the source material units.

Let $\mathbf{P}\in\mathbb{R}^{n\times3}$ contain nodal coordinates, $\mathcal{T}$ the tetrahedral connectivity, and $\mathbf{E}^{(s)}\in\mathbb{R}^{n\times1}$ the nodal Young's modulus of sample $s$. The physical prediction task is
\begin{equation}
 (\mathbf{P},\mathcal{T},\mathbf{E}^{(s)})\longmapsto
 \mathbf{y}^{(s)}=(\sigma_{\mathrm{vm},i}^{(s)})_{i=1}^{n}\in\mathbb{R}^{n\times1}.
 \label{eq:mines_task}
\end{equation}
The implemented feature matrix is $\mathbf{X}^{(s)}=[\mathcal{Z}_{P}(\mathbf{P}),\mathcal{Z}_{E}(\mathbf{E}^{(s)}),\mathbf{q}]\in\mathbb{R}^{n\times5}$, where $\mathcal{Z}$ denotes training-set standardization and $\mathbf{q}$ is the standardized log-degree feature defined below. The network predicts one scalar per node. Although the original archive contains stress-tensor components, the experiments supervise only the scalar von Mises field; displacements and stress components are not input features. Boundary conditions and load magnitudes are not separate model inputs in this fixed experimental setting.

The loader matches coordinate, modulus, connectivity, and stress files by sample identifier, and aligns modulus and stress rows to coordinate rows by node identifier. For this common mesh, the topology can be constructed once and reused. The material field, rather than connectivity, supplies the sample-dependent input.

\clearpage
\subsection{Abaqus Datasets: Geometry, Loading, and Simulation}\label{app:abaqus}

We construct two additional datasets with Abaqus/CAE 2025 and the Abaqus/Standard static solver. Their parametric scripts adapt the official connecting-lug example and the modelling workflow of the official flap-mechanism example~\citep{abaqus2025lug,abaqus2025flap}. The latter is reduced to a single perforated elbow bracket, rather than the full articulated mechanism. Both use homogeneous isotropic linear elasticity with Young's modulus $200$~GPa and Poisson's ratio $0.30$, and modified ten-node tetrahedral elements (C3D10M). Geometries are remeshed for every parameter combination. Coordinates, force, and stress use metres, newtons, and pascals in the simulation and exported data; Table~\ref{tab:simulation_parameters} expresses lengths in millimetres for readability.

\begin{table}[htbp]
\centering
\caption{Actual parameter grids, verified against the per-sample parameter records. $a:\Delta:b$ denotes values from $a$ to $b$ with increment $\Delta$. Each dataset uses the Cartesian product of its listed levels.}
\label{tab:simulation_parameters}
\small
\begin{tabular*}{\textwidth}{@{\extracolsep{\fill}}llr@{}}
\toprule
Dataset / parameter & Values & Number of levels \\
\midrule
\multicolumn{3}{l}{\textit{Connecting lug: 225 combinations}} \\
Shank length & 100 mm & 1 \\
Outer radius & $28.6:0.2:29.4$ mm & 5 \\
Pin-hole radius & 12.5 mm & 1 \\
Mounting-hole radius & $4.8,5.0,5.2$ mm & 3 \\
Thickness & $16.6:0.2:19.4$ mm & 15 \\
Global mesh seed & 6 mm & 1 \\
Total force & 20,000 N in the $-y$ direction & 1 \\
\midrule
\multicolumn{3}{l}{\textit{Elbow bracket: 288 combinations}} \\
Horizontal arm length & $160,180,200$ mm & 3 \\
Vertical arm length & $180,200,220$ mm & 3 \\
Centreline bend radius & $60,70$ mm & 2 \\
Arm width & $50,60$ mm & 2 \\
Hole radius & $9,10,11,12$ mm & 4 \\
Thickness & $18,20$ mm & 2 \\
Global mesh seed & 8 mm & 1 \\
Total force & 2,000 N & 1 \\
Force angle from $+x$ toward $+y$ & $-45^{\circ}$ & 1 \\
\bottomrule
\end{tabular*}
\end{table}

\begin{figure}[htbp]
\centering
\begin{subfigure}{\textwidth}
\centering\includegraphics[width=\linewidth]{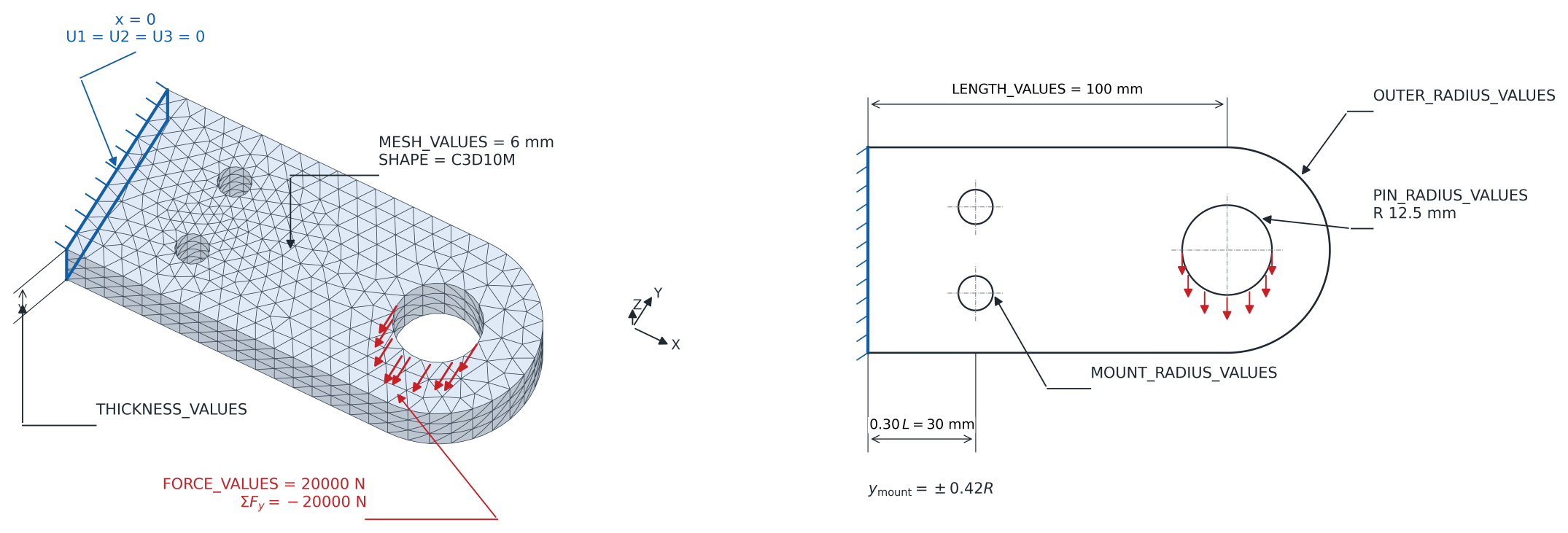}
\caption{Mesh, fixed boundary, nodal loading, and geometric parameters.}
\end{subfigure}
\par\medskip
\begin{subfigure}{\textwidth}
\centering\includegraphics[width=\linewidth]{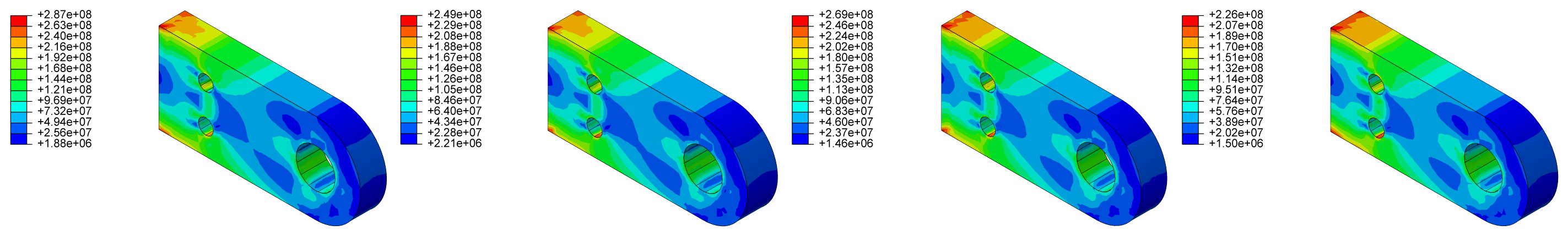}
\caption{Representative simulated von Mises stress fields for different geometries.}
\end{subfigure}
\caption{Connecting-lug dataset. The contour panels illustrate geometric variation under a fixed total load; each panel retains its own stress colour scale in Pa.}
\label{fig:lug_dataset}
\end{figure}

\paragraph{Connecting lug.}
The lug consists of a straight shank, a rounded end with a pin hole, and two smaller mounting holes. The shank face at $x=0$ is fully fixed. The prescribed resultant acts on the lower half of the pin-hole surface: if this region contains $N_{\mathrm{load}}$ mesh nodes, each receives $\mathbf{f}_i=(0,-20000/N_{\mathrm{load}},0)$~N. Thus the implemented loading is a set of concentrated nodal forces with the stated total resultant. This adapts the pressure-loaded official example using the prescribed force resultant. Varying the outer radius, mounting-hole radius, and thickness yields $5\times3\times15=225$ samples. Figure~\ref{fig:lug_dataset} shows the geometry and representative finite element stress fields.

\begin{figure}[htbp]
\centering
\begin{subfigure}{\textwidth}
\centering\includegraphics[width=\linewidth]{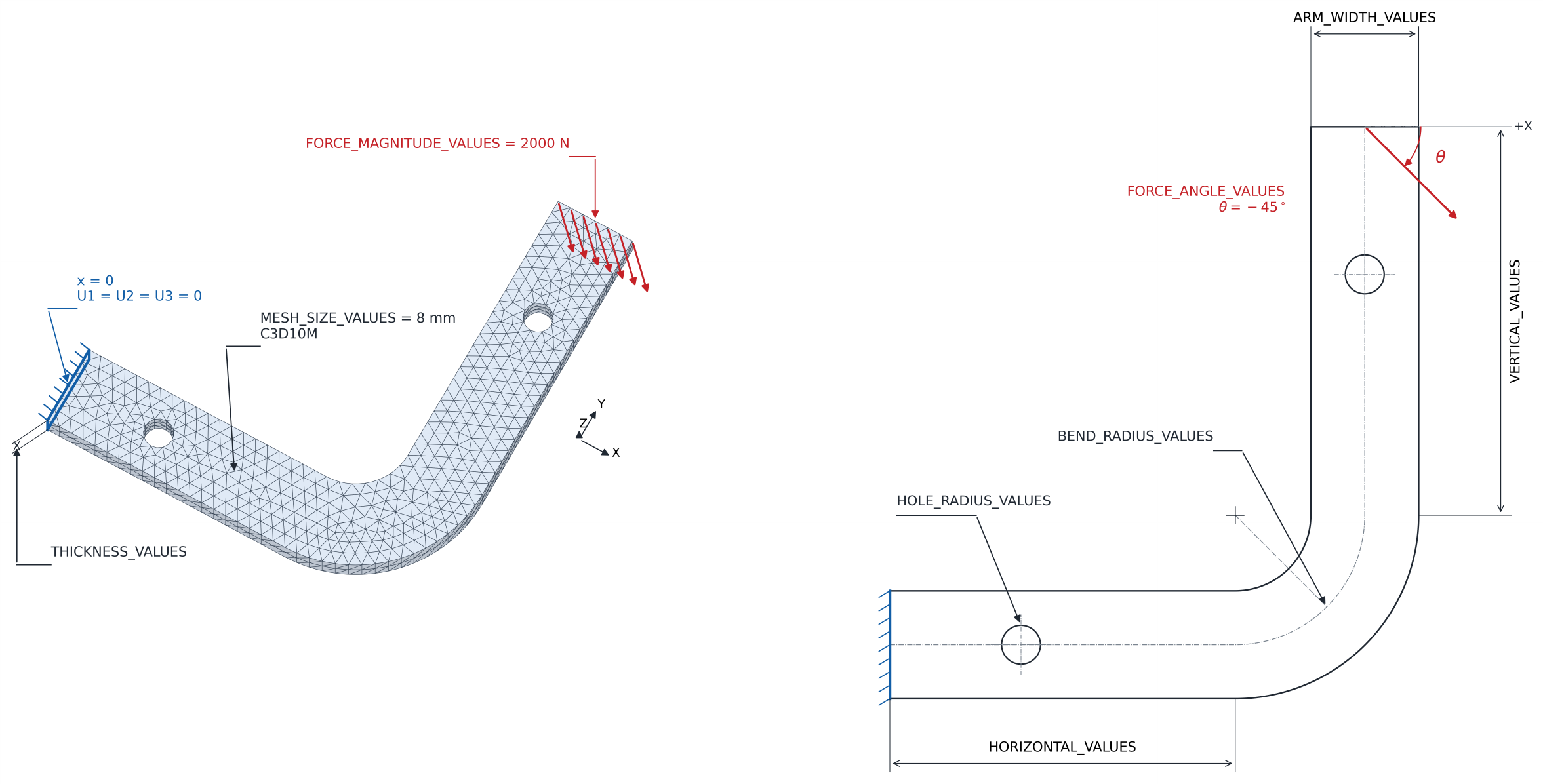}
\caption{Mesh, boundary conditions, force direction, and geometric parameters.}
\end{subfigure}
\par\medskip
\begin{subfigure}{\textwidth}
\centering\includegraphics[width=\linewidth]{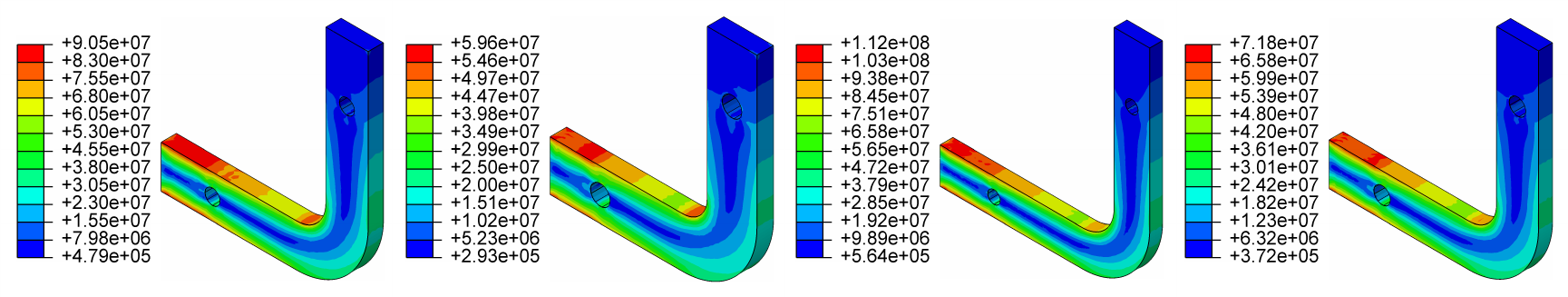}
\caption{Representative simulated von Mises stress fields for different geometries.}
\end{subfigure}
\caption{Elbow-bracket dataset. Each contour panel uses its own stress colour scale in Pa, so colours should be interpreted together with the corresponding legend.}
\label{fig:elbow_dataset}
\end{figure}

\paragraph{Elbow bracket.}
Two straight perforated arms are joined by a curved $90^{\circ}$ bend. The face at $x=0$ is fully fixed, and the force is distributed equally among nodes on the free end of the vertical arm. For $\theta=-45^{\circ}$, the nodal force is $\mathbf{f}_i=2000(\cos\theta,\sin\theta,0)/N_{\mathrm{load}}$~N. The six varied geometric parameters yield $3\times3\times2\times2\times4\times2=288$ combinations. Figure~\ref{fig:elbow_dataset} illustrates the geometry, loading, and spatially localized stresses around the bend, holes, and constrained region.

\paragraph{Stress extraction and learning task.}
The postprocessing scripts extract stress tensors at the \texttt{ELEMENT\_NODAL} positions of the selected ODB frame (the final frame by default). For a node shared by several elements, tensor components are first averaged arithmetically over the element-nodal contributions. Von Mises stress is then computed from this averaged tensor:
\begin{equation}
 \sigma_{\mathrm{vm}}=
 \sqrt{\tfrac12[(\bar\sigma_{xx}-\bar\sigma_{yy})^2+
 (\bar\sigma_{yy}-\bar\sigma_{zz})^2+(\bar\sigma_{zz}-\bar\sigma_{xx})^2]
 +3(\bar\sigma_{xy}^{\,2}+\bar\sigma_{yz}^{\,2}+\bar\sigma_{zx}^{\,2})}.
 \label{eq:abaqus_vm}
\end{equation}
This operation differs from averaging element-wise von Mises scalars. The exported stress table contains six tensor components and the scalar invariant, but training uses only the latter.

\FloatBarrier

For either custom dataset, sample $s$ defines a geometry-dependent mesh $(\mathbf{P}^{(s)},\mathcal{T}^{(s)})$ and target $\mathbf{y}^{(s)}\in\mathbb{R}^{n_s\times1}$. The physical and implemented mappings are
\begin{equation}
 (\mathbf{P}^{(s)},\mathcal{T}^{(s)})\longmapsto\mathbf{y}^{(s)},
 \qquad
 f_{\Theta}(\mathbf{X}^{(s)},\widetilde{\mathbf{A}}^{(s)})=
 \widehat{\mathbf{y}}_{\mathrm{norm}}^{(s)},\qquad
 \mathbf{X}^{(s)}=[\mathcal{Z}_{P}(\mathbf{P}^{(s)}),\mathbf{q}^{(s)}]
 \in\mathbb{R}^{n_s\times4}.
 \label{eq:custom_task}
\end{equation}
Material constants, total load, load direction, and boundary-condition rules are fixed within each dataset, so they are not supplied as additional channels. In particular, the number of loaded nodes varies with the mesh, and the simulation divides the fixed total force accordingly. These experiments assess stress prediction across geometric configurations under the prescribed physical setting, using finite element solutions as reference fields. Model comparisons on both custom datasets are reported in Appendix~\ref{app:custom_results}.

\FloatBarrier
\subsection{From Finite Element Meshes to Graphs}\label{app:graph_construction}

Each mesh node becomes a graph vertex, including the midside nodes of quadratic elements. Node labels are mapped to contiguous row indices, and target rows are aligned by label. Connectivity is used to define an undirected graph $G=(V,\mathcal{E})$; it is not treated as a dense input feature.

\paragraph{Connectivity rules.}
For Mines Paris, all distinct node pairs in each four-node tetrahedron are connected, giving its six edges. The custom-data loader uses the Abaqus C3D10M ordering: nodes 1--4 are corner nodes and 5--10 lie on edges $(1,2),(2,3),(3,1),(1,4),(2,4),(3,4)$, respectively. Each quadratic edge is split at its midside node. The twelve local segments are
\begin{equation}
\begin{split}
 \mathcal{E}_{\mathrm{tet}}=\{&(1,5),(5,2),(2,6),(6,3),(3,7),(7,1),\\
 &(1,8),(8,4),(2,9),(9,4),(3,10),(10,4)\}.
\end{split}
\end{equation}
Edges shared by elements are deduplicated globally. Each undirected edge is stored in both directions for message passing. The generic fallback in the custom-data loader connects all valid pairs for other element sizes, but every custom-data element used here has ten nodes and follows the twelve-segment rule. Thus these quadratic tetrahedra are not converted to ten-node cliques.

\paragraph{Features and graph operators.}
Let $d_i$ be the number of distinct neighbours before adding self-loops and $\ell_i=\log(1+d_i)$. The last input channel is
\begin{equation}
 q_i=\frac{\ell_i-\bar\ell}{s_{\ell}+10^{-6}},\qquad
 \bar\ell=\frac1n\sum_i\ell_i,\qquad
 s_{\ell}^{\,2}=\frac1n\sum_i(\ell_i-\bar\ell)^2.
\end{equation}
Coordinates and, for Mines Paris, Young's modulus are standardized componentwise using all nodes in the training samples only. The target is similarly standardized for optimization, with $\widehat y_i=\sigma_y\widehat y_{\mathrm{norm},i}+\mu_y$ before evaluating physical stress metrics. Any training-set standard deviation below $10^{-12}$ is replaced by one. The degree normalization is computed within each graph and does not use target values.

For LiftGCN, the binary symmetric adjacency $\mathbf{A}$ gives $\widehat{\mathbf{A}}=\mathbf{A}+\mathbf{I}$ and $\widetilde{\mathbf{A}}=\widehat{\mathbf{D}}^{-1/2}\widehat{\mathbf{A}}\widehat{\mathbf{D}}^{-1/2}$, as in Section~\ref{sec:background}. The base graph has no separately measured edge attributes. MeshGraphNets derives a four-dimensional edge feature from the difference of standardized endpoint coordinates and its Euclidean norm. Hybrid LG instead forms coordinate differences and lengths in the original coordinates, then standardizes these edge features using training-set statistics. Local-gradient evaluation uses the undirected edges without self-loops and the original, unstandardized coordinates, consistently with Eq.~(\ref{eq:hsg_metric}).

\clearpage
\section{Model Comparisons on the Custom Datasets}\label{app:custom_results}

Tables~\ref{tab:lug_results} and~\ref{tab:elbow_results} extend Section~\ref{sec:main_comparisons} to the connecting lug and elbow bracket. They report the same whole-field and local metrics as the main comparison, including all three hotspot thresholds. Prediction scores are means over 10 repetitions. The checkpoint-selection protocol is specified in Appendix~\ref{app:settings}.

\begin{table}[htbp]
\centering
\caption{Connecting-lug comparison. Prediction metrics are means over 10 repetitions. Bold marks the best score in each column, including ties. Parameters in parentheses belong to the Adv-GCN discriminator.}
\label{tab:lug_results}
\begingroup
\footnotesize
\setlength{\tabcolsep}{3.5pt}
\renewcommand{\arraystretch}{1.10}
\textbf{(a) Whole-field accuracy and computational cost}\\[3pt]
\begin{tabular*}{\textwidth}{@{\extracolsep{\fill}}lrrrr@{}}
\toprule
Model & Parameters & Latency (ms)$\downarrow$ & SNR (dB)$\uparrow$ & $R^2\uparrow$ \\
\midrule
GCN & 1512k & \textbf{0.691} & 25.021 & 0.98754 \\
Adv-GCN & 1512k (+244k) & \textbf{0.691} & 25.014 & 0.98754 \\
MeshGraphNets & 1549k & 2.383 & 26.594 & 0.99132 \\
GCNII & 1503k & 1.412 & 16.707 & 0.91574 \\
GraphCON & 1503k & 1.894 & 24.101 & 0.98463 \\
CayleyNet & 1551k & 33.694 & 28.413 & 0.99431 \\
EWGNN & 1525k & 3.855 & 18.764 & 0.94696 \\
GUMP & 1504k & 27.285 & 22.386 & 0.97719 \\
Hybrid LG & 1547k & 7.848 & 28.913 & 0.99473 \\
A-DGN & 1526k & 2.592 & 24.733 & 0.98667 \\
UniGCN & 1519k & 21.510 & 26.422 & 0.99096 \\
\midrule
LiftGCN, small & 1514k & 0.900 & 25.762 & 0.98953 \\
LiftGCN, medium & 3023k & 2.610 & 28.071 & 0.99384 \\
LiftGCN, large & 11571k & 8.870 & \textbf{29.558} & \textbf{0.99563} \\
\bottomrule
\end{tabular*}
\par\medskip
\textbf{(b) Stress-concentration reconstruction}\\[3pt]
\begin{tabular*}{\textwidth}{@{\extracolsep{\fill}}lrrrrrr@{}}
\toprule
\multirow{2}{*}{Model} & \multicolumn{3}{c}{PSNR (dB)$\uparrow$} & \multicolumn{3}{c}{HSG-NMSE$\downarrow$} \\
\cmidrule(lr){2-4}\cmidrule(l){5-7}
 & @1\% & @5\% & @10\% & @1\% & @5\% & @10\% \\
\midrule
GCN & 27.962 & 29.510 & 30.518 & 0.036797 & 0.033067 & 0.031227 \\
Adv-GCN & 27.934 & 29.487 & 30.509 & 0.035820 & 0.032702 & 0.031144 \\
MeshGraphNets & \textbf{34.056} & 34.029 & 33.867 & 0.020698 & 0.028265 & 0.029088 \\
GCNII & 19.496 & 21.695 & 22.719 & 0.498130 & 0.419740 & 0.356840 \\
GraphCON & 26.991 & 28.602 & 29.493 & 0.058009 & 0.052488 & 0.048874 \\
CayleyNet & 31.141 & 33.428 & 34.543 & \textbf{0.019631} & \textbf{0.021604} & \textbf{0.021072} \\
EWGNN & 24.741 & 25.422 & 26.082 & 0.186240 & 0.232120 & 0.228620 \\
GUMP & 26.447 & 27.979 & 28.774 & 0.059343 & 0.064130 & 0.064358 \\
Hybrid LG & 31.979 & 33.791 & 34.642 & 0.028996 & 0.028105 & 0.027341 \\
A-DGN & 28.886 & 29.853 & 30.668 & 0.032105 & 0.034136 & 0.033857 \\
UniGCN & 30.185 & 31.387 & 32.107 & 0.024199 & 0.024338 & 0.024768 \\
\midrule
LiftGCN, small & 29.517 & 30.625 & 31.367 & 0.039788 & 0.040724 & 0.040821 \\
LiftGCN, medium & 31.341 & 33.081 & 33.939 & 0.031984 & 0.037151 & 0.035538 \\
LiftGCN, large & 33.366 & \textbf{35.357} & \textbf{36.035} & 0.025049 & 0.023864 & 0.023407 \\
\bottomrule
\end{tabular*}
\endgroup
\end{table}

On the connecting lug, small LiftGCN improves SNR and peak-region PSNR over GCN at a latency of $0.900$~ms, although its HSG-NMSE is higher. Large LiftGCN attains the best SNR ($29.558$~dB), $R^2$ ($0.99563$), and PSNR at the 5\% and 10\% thresholds. It takes $8.870$~ms, compared with $21.510$~ms for UniGCN and $33.694$~ms for CayleyNet. However, MeshGraphNets has the best PSNR@1\%, CayleyNet leads all three gradient metrics, and GCN/Adv-GCN have the lowest latency. These results demonstrate a useful accuracy--latency trade-off across model sizes.

\clearpage
\begin{table}[htbp]
\centering
\caption{Elbow-bracket comparison. Prediction metrics are means over 10 repetitions. Bold marks the best score in each column, including ties. Parameters in parentheses belong to the Adv-GCN discriminator.}
\label{tab:elbow_results}
\begingroup
\footnotesize
\setlength{\tabcolsep}{3.5pt}
\renewcommand{\arraystretch}{1.10}
\textbf{(a) Whole-field accuracy and computational cost}\\[3pt]
\begin{tabular*}{\textwidth}{@{\extracolsep{\fill}}lrrrr@{}}
\toprule
Model & Parameters & Latency (ms)$\downarrow$ & SNR (dB)$\uparrow$ & $R^2\uparrow$ \\
\midrule
GCN & 1512k & \textbf{0.570} & 17.354 & 0.95714 \\
Adv-GCN & 1512k (+244k) & \textbf{0.570} & 17.175 & 0.95526 \\
MeshGraphNets & 1549k & 3.228 & 18.245 & 0.96493 \\
GCNII & 1503k & 2.383 & 8.932 & 0.70091 \\
GraphCON & 1503k & 2.663 & 16.160 & 0.94357 \\
CayleyNet & 1551k & 41.259 & 23.418 & 0.98935 \\
EWGNN & 1525k & 6.188 & 12.277 & 0.78350 \\
GUMP & 1504k & 25.150 & 11.739 & 0.84394 \\
Hybrid LG & 1547k & 13.318 & 18.945 & 0.96905 \\
A-DGN & 1526k & 4.754 & 15.492 & 0.93417 \\
UniGCN & 1519k & 27.688 & 18.997 & 0.97063 \\
\midrule
LiftGCN, small & 1514k & 2.361 & 18.246 & 0.96510 \\
LiftGCN, medium & 3023k & 4.861 & 21.216 & 0.98239 \\
LiftGCN, large & 11571k & 12.714 & \textbf{24.019} & \textbf{0.99074} \\
\bottomrule
\end{tabular*}
\par\medskip
\textbf{(b) Stress-concentration reconstruction}\\[3pt]
\begin{tabular*}{\textwidth}{@{\extracolsep{\fill}}lrrrrrr@{}}
\toprule
\multirow{2}{*}{Model} & \multicolumn{3}{c}{PSNR (dB)$\uparrow$} & \multicolumn{3}{c}{HSG-NMSE$\downarrow$} \\
\cmidrule(lr){2-4}\cmidrule(l){5-7}
 & @1\% & @5\% & @10\% & @1\% & @5\% & @10\% \\
\midrule
GCN & 24.735 & 24.915 & 25.266 & 0.124680 & 0.113800 & 0.127510 \\
Adv-GCN & 24.469 & 24.693 & 25.081 & 0.133390 & 0.120480 & 0.135120 \\
MeshGraphNets & 26.303 & 26.262 & 26.449 & 0.088613 & 0.089554 & 0.100940 \\
GCNII & 14.184 & 14.322 & 14.651 & 1.483500 & 1.731300 & 1.745500 \\
GraphCON & 23.362 & 23.781 & 24.088 & 0.255970 & 0.231220 & 0.248400 \\
CayleyNet & 28.899 & 29.158 & 29.631 & 0.054606 & 0.053158 & 0.059374 \\
EWGNN & 17.384 & 18.293 & 19.027 & 0.696530 & 0.627560 & 0.636530 \\
GUMP & 18.903 & 18.391 & 18.012 & 0.591310 & 0.604710 & 0.655690 \\
Hybrid LG & 24.311 & 25.441 & 26.196 & 0.161900 & 0.155140 & 0.158990 \\
A-DGN & 25.053 & 24.760 & 24.346 & 0.170570 & 0.192130 & 0.257930 \\
UniGCN & 26.998 & 26.656 & 27.008 & 0.053734 & 0.059160 & 0.066304 \\
\midrule
LiftGCN, small & 26.543 & 26.205 & 26.586 & 0.090990 & 0.089279 & 0.108020 \\
LiftGCN, medium & 29.036 & 28.743 & 28.998 & 0.046467 & 0.053178 & 0.062416 \\
LiftGCN, large & \textbf{29.904} & \textbf{29.985} & \textbf{30.354} & \textbf{0.038466} & \textbf{0.044894} & \textbf{0.054048} \\
\bottomrule
\end{tabular*}
\endgroup
\end{table}

On the elbow bracket, large LiftGCN leads all reported prediction metrics: SNR reaches $24.019$~dB, $R^2$ reaches $0.99074$, and HSG-NMSE@1\% falls to $0.038466$. Relative to CayleyNet, it improves SNR by $0.601$~dB and reduces HSG-NMSE@1\% by approximately $29.6\%$, with $3.25\times$ lower forward latency ($12.714$ versus $41.259$~ms). Medium LiftGCN also improves every prediction metric over UniGCN at $4.861$ versus $27.688$~ms. GCN remains the fastest model. Across both datasets, greater LiftGCN capacity improves every reported prediction metric, with parameter counts and latency documenting the computational cost of scaling.

\clearpage
\section{Experimental Settings}\label{app:settings}

\subsection{Shared Training and Evaluation Protocol}\label{app:training_protocol}

Table~\ref{tab:shared_settings} summarizes the common training protocol in the experiment scripts, with the observed model-specific exceptions explicitly noted. Each repetition shuffles sample identifiers with a NumPy generator seeded by $42+r$, for $r=0,\ldots,9$, and assigns the first $\lfloor0.8N\rfloor$ samples to training. The remaining samples form the evaluation partition. The split is at the simulation level, not at the node level. Models using the same dataset and repetition seed receive the same partition. Normalization statistics are recomputed from the training partition for each repetition.

\begin{table}[htbp]
\centering
\caption{Training settings shared across the three datasets, as implemented in the supplied scripts. Model-specific configurations are listed in Tables~\ref{tab:mines_configs} and~\ref{tab:custom_configs}.}
\label{tab:shared_settings}
\small
\begin{tabular}{@{}p{0.29\textwidth}p{0.66\textwidth}@{}}
\toprule
Setting & Value / procedure \\
\midrule
Repetitions and seeds & 10 runs; seeds 42--51 for Python, NumPy, and PyTorch \\
Training/evaluation ratio & 80\%/20\%; counts are given in Table~\ref{tab:dataset_statistics} \\
Batch and sample order & One complete mesh per optimizer update; shuffle training meshes each epoch using seed plus epoch index \\
Training duration & 50 epochs, without early termination \\
Predictor optimizer & AdamW, learning rate $2\times10^{-3}$, weight decay $10^{-5}$ \\
Learning-rate schedule & Cosine annealing, $T_{\max}=50$, minimum learning rate 0 \\
Gradient clipping & Global parameter-gradient norm bounded by 1.0 \\
Regression objective & Mean squared error over nodes in standardized stress units \\
Dropout & 0.10 \\
Adv-GCN exception & Predictor additionally uses an adversarial loss; discriminator optimized separately with Adam \\
Checkpoint choice & Lowest evaluation-partition MSE over 50 epochs; first minimum retained on a tie \\
Final scores & Restore the chosen predictor checkpoint, undo target standardization, evaluate, then average scores over 10 runs \\
Inference timing & CUDA events; 20 warm-up and 100 timed forward calls per profiled graph shape, outside training \\
\bottomrule
\end{tabular}
\end{table}

\paragraph{Checkpoint selection.}
After each epoch, we compute the test-set MSE in standardized stress units, averaging the node-mean MSE over test meshes. For each repetition, we retain the checkpoint with the lowest test-set MSE across 50 epochs and restore it to compute the reported metrics. This procedure is used for all three datasets.

\paragraph{Inference timing.}
Latency is the average forward inference time per sample on the NVIDIA RTX PRO 6000. The benchmark places graph tensors on the GPU before timing, switches the predictor to evaluation mode, and measures forward calls with synchronized CUDA events. CSV loading, graph construction, host-to-device transfer, normalization, and metric calculation are excluded. The LiftGCN scripts cache timing measurements by the tuple (node count, input dimension, adjacency nonzeros) within each partition, profiling one representative of each shape. Each profiled forward call processes one complete mesh, and the timings are averaged across the profiled shapes and calls. Adv-GCN uses only its predictor at inference; its discriminator parameters are reported separately.

\subsection{Model-Specific Configurations}\label{app:model_configs}

Tables~\ref{tab:mines_configs} and~\ref{tab:custom_configs} specify the model configurations used for Mines Paris and the two custom datasets, respectively. Parameter names follow the implementation. Connecting lug and elbow bracket use the same model configurations. Shared optimization settings are given in Table~\ref{tab:shared_settings}.

\begin{table}[htbp]
\centering
\caption{Model configurations for Mines Paris.}
\label{tab:mines_configs}
\small
\renewcommand{\arraystretch}{1.15}
\begin{tabular}{@{}p{0.23\textwidth}p{0.73\textwidth}@{}}
\toprule
Model & Model-specific hyperparameters \\
\midrule
GCN & \raggedright \texttt{hidden\_dim=270}, \texttt{layers=6} \tabularnewline
Adv-GCN & \raggedright \texttt{hidden\_dim=270}, \texttt{layers=6}, \texttt{adv\_beta=0.1}, \texttt{disc\_hidden\_dim=128}, \texttt{disc\_layers=4} \tabularnewline
MeshGraphNets & \raggedright \texttt{mgn\_latent\_size=64}, \texttt{mgn\_message\_passing\_steps=10} \tabularnewline
GCNII & \raggedright \texttt{hidden\_dim=270}, \texttt{layers=6}, \texttt{gcnii\_alpha=0.1}, \texttt{gcnii\_lambda=0.50} \tabularnewline
GraphCON & \raggedright \texttt{hidden\_dim=270}, \texttt{layers=6}, \texttt{graphcon\_dt=1.0}, \texttt{graphcon\_alpha=1.0}, \texttt{graphcon\_gamma=1.0} \tabularnewline
CayleyNet & \raggedright \texttt{hidden\_dim=110}, \texttt{layers=6}, \texttt{cayley\_order=3}, \texttt{cayley\_jacobi\_iters=5}, \texttt{cayley\_h\_init=1.0} \tabularnewline
EWGNN & \raggedright \texttt{hidden\_dim=195}, \texttt{layers=4}, \texttt{ewgnn\_alpha=0.5}, \texttt{ewgnn\_eta=1.0} \tabularnewline
GUMP & \raggedright \texttt{hidden\_dim=240}, \texttt{layers=4}, \texttt{base\_layers=1}, \texttt{gump\_attn\_dim=32}, \texttt{gump\_ns\_iters=10} \tabularnewline
Hybrid LG & \raggedright \texttt{hidden\_dim=90}, \texttt{layers=6}, \texttt{attention\_frequency=3} \tabularnewline
A-DGN & \raggedright \texttt{hidden\_dim=400}, \texttt{layers=6}, \texttt{adgn\_epsilon=0.1}, \texttt{adgn\_gamma=0.1} \tabularnewline
UniGCN & \raggedright \texttt{hidden\_dim=235}, \texttt{layers=4}, \texttt{taylor\_order=10}, \texttt{uniconv\_time=1.0} \tabularnewline
LiftGCN, small & \raggedright \texttt{hidden\_dim=330}, \texttt{layers=4} \tabularnewline
LiftGCN, medium & \raggedright \texttt{hidden\_dim=640}, \texttt{layers=8} \tabularnewline
LiftGCN, large & \raggedright \texttt{hidden\_dim=1280}, \texttt{layers=10} \tabularnewline
\bottomrule
\end{tabular}
\end{table}
\FloatBarrier

\begin{table}[htbp]
\centering
\caption{Shared model configurations for connecting lug and elbow bracket.}
\label{tab:custom_configs}
\small
\renewcommand{\arraystretch}{1.15}
\begin{tabular}{@{}p{0.23\textwidth}p{0.73\textwidth}@{}}
\toprule
Model & Model-specific hyperparameters \\
\midrule
GCN & \raggedright \texttt{hidden\_dim=500}, \texttt{layers=6} \tabularnewline
Adv-GCN & \raggedright \texttt{hidden\_dim=500}, \texttt{layers=6}, \texttt{adv\_beta=0.1}, \texttt{disc\_hidden\_dim=200}, \texttt{disc\_layers=6} \tabularnewline
MeshGraphNets & \raggedright \texttt{mgn\_latent\_size=160}, \texttt{mgn\_message\_passing\_steps=6} \tabularnewline
GCNII & \raggedright \texttt{hidden\_dim=500}, \texttt{layers=6}, \texttt{gcnii\_alpha=0.1}, \texttt{gcnii\_lambda=0.50} \tabularnewline
GraphCON & \raggedright \texttt{hidden\_dim=500}, \texttt{layers=6}, \texttt{graphcon\_dt=1.0}, \texttt{graphcon\_alpha=1.0}, \texttt{graphcon\_gamma=1.0} \tabularnewline
CayleyNet & \raggedright \texttt{hidden\_dim=210}, \texttt{layers=6}, \texttt{cayley\_order=3}, \texttt{cayley\_jacobi\_iters=5}, \texttt{cayley\_h\_init=1.0} \tabularnewline
EWGNN & \raggedright \texttt{hidden\_dim=290}, \texttt{layers=6}, \texttt{ewgnn\_alpha=0.5}, \texttt{ewgnn\_eta=1.0} \tabularnewline
GUMP & \raggedright \texttt{hidden\_dim=380}, \texttt{layers=6}, \texttt{base\_layers=2}, \texttt{gump\_attn\_dim=32}, \texttt{gump\_ns\_iters=10} \tabularnewline
Hybrid LG & \raggedright \texttt{hidden\_dim=170}, \texttt{layers=6}, \texttt{attention\_frequency=3} \tabularnewline
A-DGN & \raggedright \texttt{hidden\_dim=780}, \texttt{layers=6}, \texttt{adgn\_epsilon=0.1}, \texttt{adgn\_gamma=0.1} \tabularnewline
UniGCN & \raggedright \texttt{hidden\_dim=355}, \texttt{layers=6}, \texttt{taylor\_order=10}, \texttt{uniconv\_time=1.0} \tabularnewline
LiftGCN, small & \raggedright \texttt{hidden\_dim=500}, \texttt{layers=6} \tabularnewline
LiftGCN, medium & \raggedright \texttt{hidden\_dim=500}, \texttt{layers=12} \tabularnewline
LiftGCN, large & \raggedright \texttt{hidden\_dim=800}, \texttt{layers=18} \tabularnewline
\bottomrule
\end{tabular}
\end{table}
\FloatBarrier

\end{document}